\documentclass[11pt]{article}
\usepackage{amsfonts}

\usepackage[preprint]{acl}

\usepackage{times}
\usepackage{latexsym}

\usepackage[T1]{fontenc}

\usepackage[utf8]{inputenc}

\usepackage{microtype}

\usepackage{inconsolata}

\usepackage{graphicx}
\usepackage{amsmath}
\usepackage{xcolor}
\usepackage{enumitem}
\usepackage{tcolorbox}
\usepackage{float} 

\usepackage[table]{xcolor}
\usepackage{tcolorbox}
\tcbuselibrary{skins, breakable}
\usepackage{enumitem}
\usepackage{multirow}
\usepackage{booktabs}
\usepackage{multirow}
\usepackage{graphicx}
\usepackage{makecell}
\usepackage{arydshln}
\usepackage{booktabs}
\usepackage{multirow}
\usepackage{makecell}
\usepackage{pdflscape}
\usepackage{longtable}
\usepackage{arydshln}
\usepackage{array}

\definecolor{darkyellow}{RGB}{204,153,0}
\definecolor{lightyellow}{RGB}{255,248,220}

\definecolor{headergray}{RGB}{50, 50, 50}
\definecolor{lightgrey}{gray}{0.95}
\definecolor{darkred}{RGB}{120, 0, 0}
\definecolor{darkblue}{RGB}{0, 50, 100}
\colorlet{lightred}{red!2}
\colorlet{lightblue}{cyan!2}
\definecolor{lightyellow}{RGB}{255,255,240}
\newtcolorbox{datasetbox}[1]{
    colback=white,
    colframe=headergray,
    fontupper=\ttfamily\footnotesize,
    arc=3pt,
    outer arc=3pt,
    boxrule=1pt,
    left=10pt,
    right=10pt,
    top=10pt,
    bottom=10pt,
    enhanced,
    attach boxed title to top left={yshift=-2mm, xshift=5pt},
    boxed title style={
        colback=headergray,
        sharp corners,
        boxrule=0pt,
        top=2pt,
        bottom=2pt,
        left=5pt,
        right=20pt
    },
    title={\bfseries\normalsize #1}
}

\newtcolorbox{casebox}[2]{
    enhanced,
    fontupper=\footnotesize,
    colframe=#1,
    colback=#2,
    arc=0mm,
    boxrule=0.8pt,
    left=5pt,
    right=5pt,
    top=5pt,
    bottom=5pt,
    boxsep=0pt,
    segmentation style={draw=#1, solid, line width=0.5pt}
}

\newtcolorbox{outerbox}{
    enhanced,
    fontupper=\footnotesize,
    colframe=black,
    colback=white,
    arc=0mm,
    boxrule=1pt,
    left=2pt,
    right=2pt,
    top=2pt,
    bottom=2pt
}

\usepackage{booktabs}
\usepackage{multirow, booktabs, xcolor, colortbl}

\title{Ask or Answer: A Decision Framework for Multi-Turn Health Misinformation Intervention}

\author{
  \textbf{Xiaoying Song\textsuperscript{1}} \quad
  \textbf{Anirban Saha Anik\textsuperscript{1}} \quad
  \textbf{Jinyu Liu\textsuperscript{1}} \\
  \textbf{Qitao Tan\textsuperscript{2}} \quad
  \textbf{Geng Yuan\textsuperscript{2}} \quad
  \textbf{Lingzi Hong\textsuperscript{1}}
\\
  \textsuperscript{1} University of North Texas
\\
  \textsuperscript{2} University of Georgia
\\
  \small{
  \{xiaoyingsong, anirbansahaanik, jinyuliu\}@my.unt.edu,
  lingzi.hong@unt.edu
  }
\\
  \small{
  \{qitaotan, geng.yuan\}@uga.edu
  }
}

\begin{document}
\maketitle
\begin{abstract}
Correcting health misinformation in dialogue requires more than producing a factual rebuttal: users differ in what they know, what they believe, and what they need to hear, so an effective intervention often depends on first asking the right clarifying question. Yet existing methods either respond immediately or probe indiscriminately, treating clarification as either unnecessary or always beneficial. We propose \textbf{Reward-Optimized Probe-and-Respond (RO-PnR)}, a framework that learns \textit{when asking is worth its cost}. At each turn, RO-PnR chooses between probing for more information and committing to a final correction, guided by a turn-level reward that weighs the expected gain from probing against its interaction cost. To capture how user heterogeneity affects probing value, we model each simulated user with a latent state along health literacy and belief commitment. Experiment shows that RO-PnR achieves the highest cost-adjusted utility across three health-misinformation datasets and three base models, using $\sim$30\% fewer turns than always-probe baselines.
\end{abstract}

\section{Introduction}

\begin{figure}
    \centering
    \includegraphics[width=0.75\linewidth]{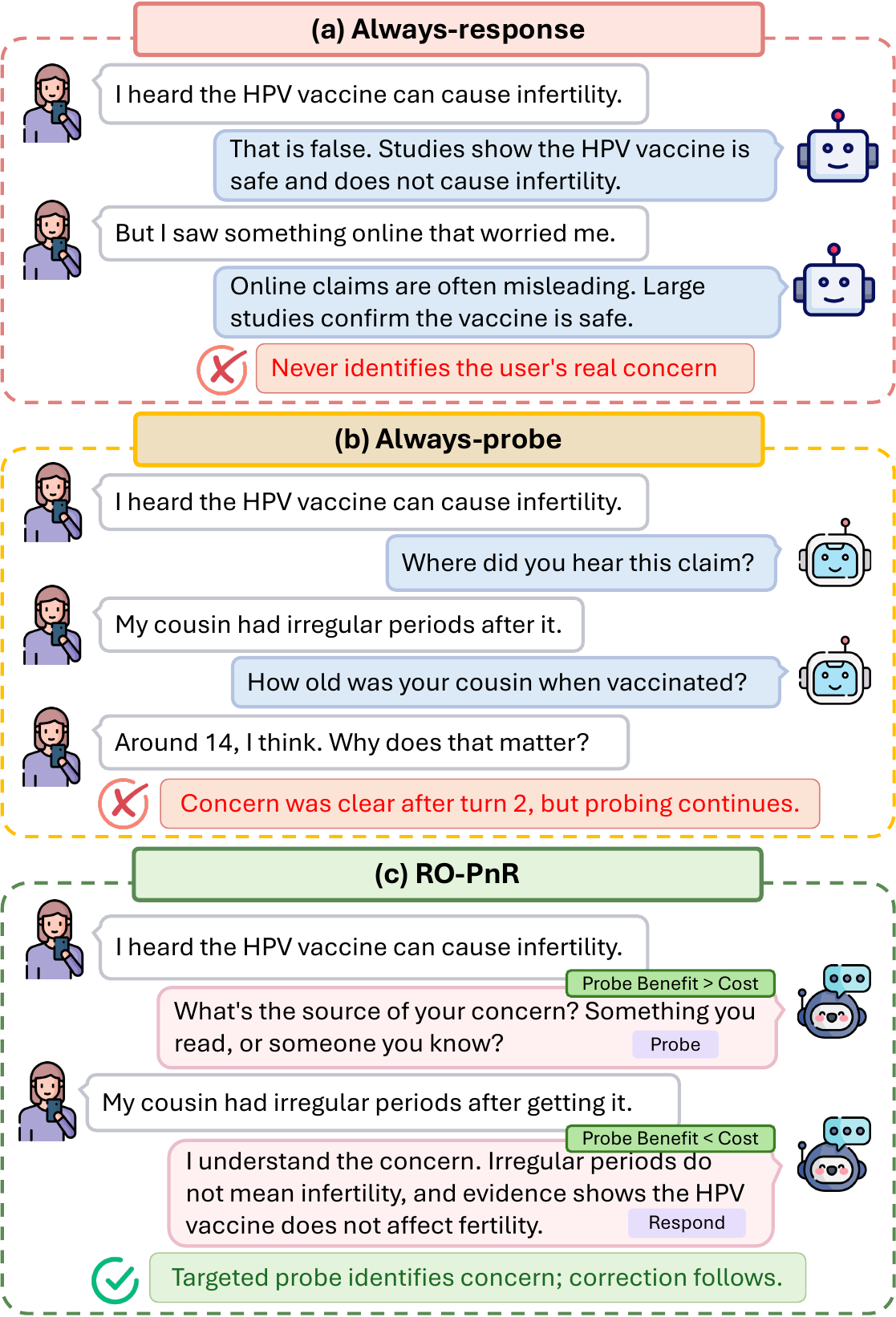}
    \caption{Different strategies on the same misinformation post. RO-PnR weighs benefit vs. cost before acting.
    }
    \label{fig:placeholder}
\end{figure}

Health misinformation distorts public health, undermining vaccine uptake and adherence to public-health guidance~\cite{van2022misinformation}, and dialogue-based agents are an increasingly promising channel for correcting it~\cite{karinshak2023working}. Effective correction depends on more than producing accurate facts; it depends on what the specific user knows, what they believe, and why they believe it~\cite{ecker2022psychological}.

Prior work on health misinformation correction has focused on the response itself: generating evidence-grounded rebuttals~\cite{yue2024evidence,anik2025multi} or tailoring them to user characteristics such as health literacy~\cite{song2025speaking, peng2024rescuing}. These methods improve the quality of individual responses, but they commit to a reply based on the original post alone, without engaging the user to surface the underlying concern. A separate line of work on multi-turn dialogue uses clarification questions to handle ambiguity~\cite{chen2024learning, wu2025collabllm}, with recent decision-theoretic methods weighing the value of asking against its cost~\cite{dong2026value, NEURIPS2024_32b80425}. 
However, these approaches typically ask follow-up questions to clarify users’ ambiguous queries for task completion, without considering users' underlying state. In health misinformation intervention, users differ in their knowledge and beliefs, so the effects of asking versus answering may vary across users. 
This makes correction a decision problem, not just a generation problem. 

At every interaction turn, the agent must decide whether to respond now with the information it has or ask the user a clarifying question first~\cite{chen2024learning, zhang2025clarify}. 
Responding too eagerly commits to a generic correction before the user's real concern surfaces (Figure 1a). Asking too eagerly, probing at every opportunity, wastes the user's attention and adds little when the context is already sufficient (Figure 1b). The right behavior depends on whether the expected benefit of one more question outweighs the cost of asking it~\cite{dong2026value}.

The trade-off is sharpened by user heterogeneity. Users vary in health literacy: what they can understand and reason about~\cite{nutbeam2000health, berkman2011low}, and in belief commitment~\cite{ecker2022psychological, wittenberg2020misinformation}: how firmly they hold the misconception. A low-literacy user who is open to revision may need a single targeted probe; a high-literacy user who already articulated their concern may need none~\cite{song2025speaking}; a strong-believer user may need careful probing precisely to gather the context that will make a correction non-confrontational~\cite{swire2022backfire, lewandowsky2012misinformation}. The value of asking, therefore, is itself user-dependent. Agents that don't model this heterogeneity will systematically over- or under-probe.

We propose Reward-Optimized Probe-and-Respond (RO-PnR), a framework that learns when asking is worth its cost. At each turn, RO-PnR chooses between probing and responding based on a turn-level decision reward that weighs the expected gain from probing against the cost of one more question. Because the value of probing depends on hidden user characteristics, we model each simulated user with a latent state along health literacy and belief commitment, and train the policy on turn-level decisions. As illustrated in Figure 1c, RO-PnR learns to ask one well-targeted question when context is missing, and to commit immediately when it is not.
We validate this approach across three health-misinformation datasets and three base models. RO-PnR achieves the highest cost-adjusted utility while using ~30\% fewer turns than always-probe baselines, with the largest gains on low-literacy and strongly-committed users, exactly the slices where the probing decision matters most.

\textbf{Our contributions are:}
(a) We frame health misinformation intervention as a probe-or-respond decision problem grounded in domain-specific user dynamics, moving beyond the single-turn, one-size-fits-all paradigm of prior counterspeech work.
(b) We introduce \textbf{RO-PnR}, a decision-theoretic policy tailored to health misinformation intervention that learns when asking is worth its cost, accounting for user heterogeneity along health literacy and belief commitment.


\section{Related Work}
\subsection{Health Misinformation Intervention}
Research on health misinformation intervention has progressed from generic factual rebuttals to retrieval-augmented generation that grounds corrections in scientific evidence~\cite{yue2024evidence, he2023reinforcement} and audience-aware approaches that tailor responses to the user's health literacy or belief commitment~\cite{song2025speaking, peng2024rescuing, anik2025multi}. 
These methods improve the quality of individual responses but remain single-turn: the agent commits to a reply based on the misinformation content, without engaging the user to surface the underlying concern or adapting the response as new information emerges. This is particularly limiting in heterogeneous user settings, where the right intervention depends on context~\cite{joseph2025decide}.



\subsection{Decision-Theoretic Probing in Dialogue}
Recent work treats clarification as an active decision rather than a default behavior. Some methods train models to choose between asking and answering when a request is ambiguous~\cite{chen2024learning, zhang2025clarify}, others use fixed-schedule probing that asks a predetermined set of questions before responding~\cite{fu2025first}, and decision-theoretic approaches weigh the benefit of asking against its communication cost~\cite{dong2026value, NEURIPS2024_32b80425}. Closer to our setting, \citet{wu2025collabllm} introduces multi-turn-aware rewards that estimate the long-term value of a response via simulated future conversations, and \citet{wan2025enhancing} adds a curiosity reward to reduce uncertainty about users' latent state. 
We build on this direction by combining forward-looking reward estimation with an explicit interaction cost, and by grounding the latent user state in domain-specific dimensions (health literacy and belief commitment) rather than a generic user variable.

\subsection{User Heterogeneity Modeling}
Recent studies adapt dialogue agents to user-specific traits, either by conditioning responses on explicit profiles~\cite{salemi2024lamp} or by inferring latent representations from interaction~\cite{ wang2025know}. Research has shown that responses tailored to user literacy level are more effective~\cite{song2025speaking, peng2024rescuing}. 
Our work differs: (1) We treat the user state as a hidden variable that drives the agent's \emph{decision to probe} and shapes its response; (2) We model two coupled user-state dimensions, health literacy and belief commitment, which jointly shape misinformation susceptibility~\cite{ecker2022psychological, nan2022why} and are key to health misinformation intervention. 
\begin{figure*}
    \centering
    \includegraphics[width=0.8\linewidth]{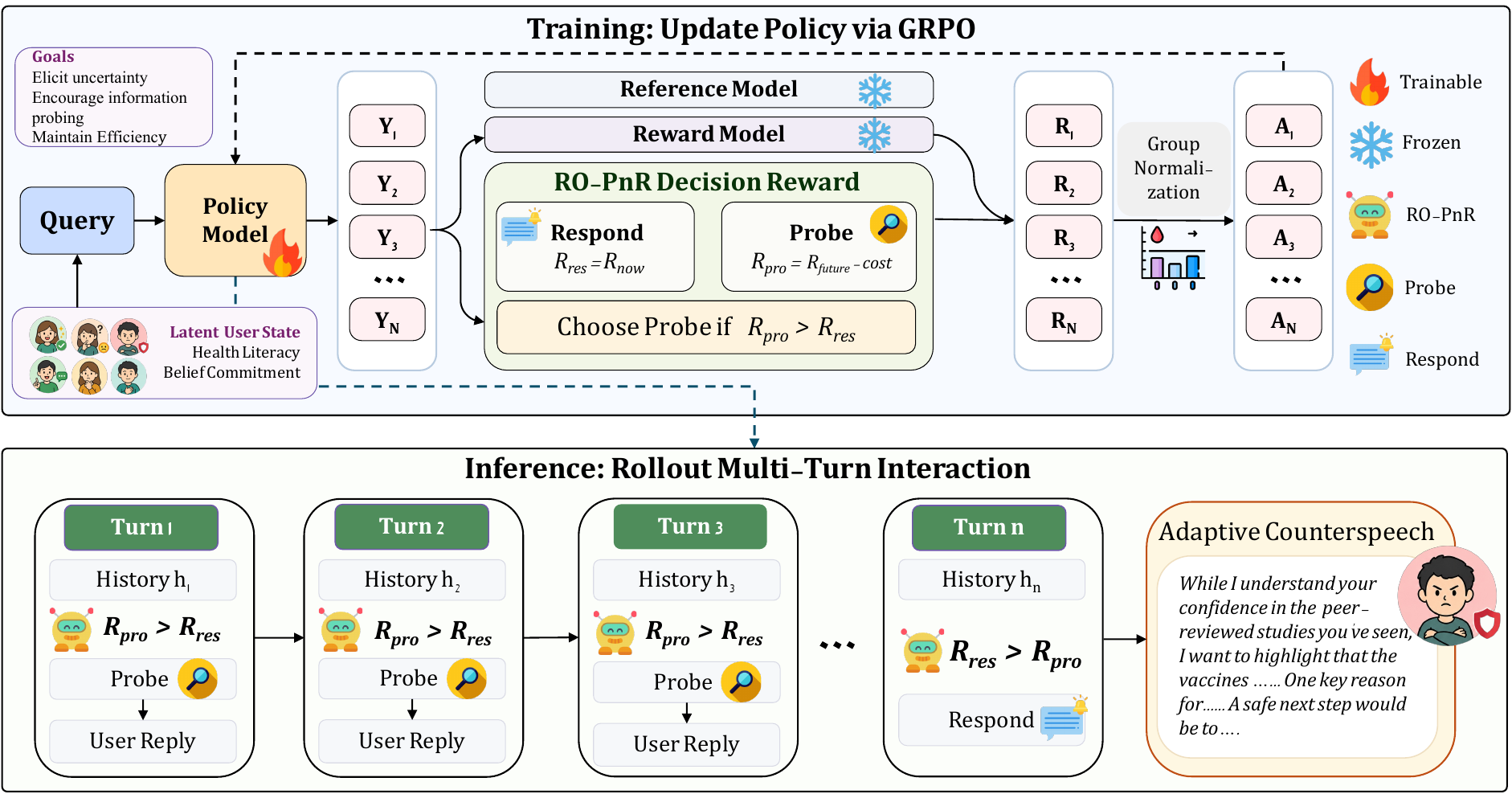}
  \caption{\textbf{Overview of the RO-PnR framework.} \textit{Top (training):} the policy generates candidate actions conditioned on the query and hidden user state; a frozen reference and reward model score the dialogue, and the reward compares \textsc{Respond} (current quality) against \textsc{Probe} (expected future quality minus cost). Group-normalized advantages then update the policy via GRPO. \textit{Bottom (inference):} the policy interacts with the user over multiple turns, deciding at each step whether to probe or produce a final adaptive response.}
    \label{fig:placeholder}
\end{figure*}

\section{Methodology}

\subsection{Task Definition}
We formulate the problem as a multi-turn health misinformation intervention task under hidden user heterogeneity. Given a health-misinformation post $x$, the agent interacts with a user over a short dialogue and follows a RO-PnR paradigm. At each turn, the agent may either \emph{probe} to elicit missing users' concern or \emph{respond} with a final correction.

\subsection{RO-PnR Policy}
The policy casts intervention as a sequential decision problem: the agent must decide not only \emph{when} to probe~\cite{zhu2025ask}, but also \emph{when} to stop gathering information and commit to a final response~\cite {dong2026value}.

Let the dialogue history at turn $t$ be
\[
h_t = \{x,(u_1,a_1),\ldots,(u_{t-1},a_{t-1}),u_t\},
\]
where $u_t$ denotes the current user utterance. Conditioned on $h_t$, the agent selects an action
\[
a_t \sim \pi(\cdot \mid h_t),
\]
where $\pi$ denotes the RO-PnR policy and
\[
a_t \in \{\textsc{Probe},\textsc{Respond}\}.
\]
At each turn, the agent must decide whether the current information in $h_t$ is sufficient to generate a final adaptive response, or whether asking one additional clarification question is likely to yield enough benefit to justify its interaction cost. Accordingly, the policy chooses \textsc{Probe} only when the expected value of acquiring additional user information outweighs the cost of continuing the dialogue; otherwise, it chooses \textsc{Respond}.

\subsection{Latent User Simulation}
The RO-PnR policy operates over the observed dialogue history $h_t$. However, the evolution of this history depends on how different users respond to the intervention. To capture such hidden heterogeneity, we model each simulated user with a latent state $z \in \mathcal{Z}$ that governs how they react to the agent's actions.

Prior work identifies two dimensions as central to health misinformation intervention: \emph{health literacy}~\cite{berkman2011low, sorensen_health_2012, nan2022why} and \emph{belief commitment}~\cite{ecker_psychological_2022, Walter03072018, lewandowsky2012misinformation}. Health literacy refers to a user's ability to understand and process health-related information, which directly affects how they interpret evidence and make health decisions~\cite{ogbadu2026information}. Belief commitment reflects the strength with which a user holds a misinformation-related belief, as well as their resistance to corrective information~\cite{siebert2023effective, wittenberg2020misinformation}; 
These two dimensions are tightly coupled: A user’s response to correction depends on both their ability to understand the evidence and their willingness to revise prior beliefs. Effective intervention must therefore account for both comprehension capacity and openness to belief revision.

Motivated by these findings, we model each user along these two dimensions and define the latent state space as
\[
\mathcal{Z} = \mathcal{L} \times \mathcal{B},
\]
where $\mathcal{L}$ denotes health literacy and $\mathcal{B}$ denotes belief commitment. Following \citet{nutbeam2000health}, we discretize health literacy as
\[
\mathcal{L} = \{\textit{functional},\ \textit{interactive},\ \textit{critical}\},
\]
corresponding to increasing capacity to understand, apply, and critically evaluate health information. Since there is no established categorization of belief commitment, we draw on prior work~\cite{siebert2023effective, wittenberg2020misinformation} to define
\[
\mathcal{B} = \{\textit{strong},\ \textit{hesitant},\ \textit{open}\},
\]
reflecting decreasing resistance to belief revision. Together, $(L, B) \in \mathcal{Z}$ provide a compact characterization of user heterogeneity that supports adaptive intervention.
We provide detailed explanations of user modeling in Appendix~\ref{Appendix: User Modeling Definition} and Figure~\ref{fig:9_profile}.

\subsection{Reward Design}
\label{Reward Design}

We train the RO-PnR policy with a turn-level decision reward~\cite{zhou2024archer, gao2024regressing} built in two layers, each addressing a distinct aspect of the probe-or-respond trade-off.

\textbf{(1) Forward-looking value.}
The value of asking is not in the question itself, but in the better intervention it enables later in the dialogue. A reward that scores only the next utterance misses this: a probe may add little immediately but unlock a substantially long horizon benefit, like better final response~\cite{wu2025collabllm}. We therefore compare the response quality available now against the expected quality at conversation end after probing:
\begin{align*}
r_t(\textsc{Respond})    &= R(h_t), \\
r_t(\textsc{Probe}) &= \mathbb{E}\big[\, R(h_{\mathrm{end}}) \mid h_t,\, \textsc{Probe}\,\big],
\end{align*}
where $R(\cdot)$ scores a dialogue \textit{Quality} using the average of audience alignment, personalized grounding, and tailored actionability considering dialogue history and response, and $h_{\mathrm{end}}$ is the state at which the agent eventually responds. 
(See Section~\ref{Evaluation Metrics} for \textit{Quality} evaluation details.)

\textbf{(2) Interaction cost.}
A forward-looking gain alone treats every probe as free, which would incentivize the agent to ask whenever any improvement is expected, even arbitrarily small ones. In practice, however, every clarification question imposes a real cost~\cite{dong2026value}. An agent that ignores this cost will over-probe, degrading user experience even when the marginal information gain is negligible. We therefore subtract a per-question cost from the probe action that scales with the number of clarifications already asked, referring to ~\cite{dong2026value}:
\[
r_t(\textsc{Probe}) = R_{\mathrm{future}} - c \cdot n_t,
\]
where $c > 0$ is the per-question cost and $n_t$ is the number of clarification questions asked up to turn $t$. 
This increases the cost of probing as the dialogue progresses, requiring each additional question to yield greater expected benefit to balance cumulative cost.
We further experiment with different cost selections in Section \ref{Ablation Study}.

\paragraph{Putting it together.}
At each turn, the agent picks the action with the higher reward, choosing \textsc{Probe} when
\[
R_{\mathrm{future}}(h_t) - R(h_t) \;>\; c \cdot n_t,
\]
and \textsc{Respond} otherwise. Intuitively, the agent probes only when the expected long-horizon improvement from one more clarification outweighs the cumulative interaction cost. Early probes face a low bar, but later probes must justify a higher cost, encouraging the agent to gather information quickly and commit once further probing offers diminishing returns.

\subsection{RO-PnR Optimization}
We train the RO-PnR policy on a dataset of \emph{turn-level decisions} rather than full dialogues, so that supervision aligns with the policy's per-step probe-or-respond choice. At each decision state extracted from our multi-turn rollouts, we build paired supervision: a \textsc{Respond} record scored by the immediate response quality, and a \textsc{Probe} record scored by the best continuation quality reachable after one further question. Pairing the two alternatives at the same state lets the policy learn the local trade-off directly. Construction details (rollout sources, grouping by post and profile, filtering) are deferred to Appendix~\ref{Appendix: Training Dataset Construction}.

We first warm-start the policy with supervised fine-tuning on the best resolved trajectory in each group, then optimize it with Group Relative Policy Optimization (GRPO)~\cite{shao2024deepseekmath}. GRPO suits our binary turn-level setting because it derives advantages by comparing candidate actions within the same decision group, eliminating the need for a separate value network. Each group is scored with the reward from Section~\ref{Reward Design}, normalized into group-relative advantages, and used to update the policy. Because supervision operates directly at the turn level, the policy learns \emph{when} to probe and \emph{when} to respond at each step.

\section{Experiment Setup}
\definecolor{ourrow}{HTML}{E8F1FB}    
\definecolor{utilcol}{HTML}{FFF4D6}  
\newcommand{\ours}[1]{\textbf{#1}\textsuperscript{$\star$}}  

\begin{table*}[t]
\centering
\setlength{\tabcolsep}{3pt}
\renewcommand{\arraystretch}{1.1}
\resizebox{\textwidth}{!}{%
\begin{tabular}{ll|ccccccc|ccccccc|ccccccc}
\toprule
\multirow{2}{*}{Dataset} & \multirow{2}{*}{Method} & \multicolumn{7}{c|}{\textbf{Llama-3.1-8B-Instruct}} & \multicolumn{7}{c|}{\textbf{Qwen3-8B}} & \multicolumn{7}{c}{\textbf{Gemma-4-E4B-it}} \\
\cmidrule(lr){3-9} \cmidrule(lr){10-16} \cmidrule(lr){17-23}
& & AA & PG & TA & Quality & FER$\downarrow$ & Turns$\downarrow$ & \textbf{Utility} & AA & PG & TA & Quality & FER$\downarrow$ & Turns$\downarrow$ & \textbf{Utility} & AA & PG & TA & Quality & FER$\downarrow$ & Turns$\downarrow$ & \textbf{Utility} \\
\midrule
\multirow{6}{*}{\textbf{CounterHealth}} & Single-Turn & 0.58 & 0.53 & 0.61 & 0.57 & 0.18 & 5.00 & \cellcolor{utilcol}0.56 & 0.58 & 0.52 & 0.61 & 0.57 & 0.15 & 5.00 & \cellcolor{utilcol}0.56 & 0.64 & 0.59 & 0.67 & 0.63 & 0.04 & 5.00 & \cellcolor{utilcol}0.62 \\
 & Fixed-Q & 0.72 & 0.65 & 0.76 & 0.71 & 0.11 & 5.00 & \cellcolor{utilcol}0.70 & 0.71 & 0.64 & 0.73 & 0.69 & 0.08 & 5.00 & \cellcolor{utilcol}0.68 & \textbf{0.73} & 0.66 & 0.74 & 0.71 & 0.02 & 5.00 & \cellcolor{utilcol}0.70 \\
 & Reactive & 0.72 & 0.64 & 0.76 & 0.70 & 0.11 & 5.00 & \cellcolor{utilcol}0.69 & 0.71 & 0.64 & 0.72 & 0.69 & 0.09 & 5.00 & \cellcolor{utilcol}0.68 & 0.72 & 0.65 & 0.74 & 0.71 & \textbf{0.02} & 3.92 & \cellcolor{utilcol}0.70 \\
 & Confidence & 0.71 & 0.64 & 0.74 & 0.69 & 0.13 & \textbf{1.26} & \cellcolor{utilcol}0.69 & 0.67 & 0.61 & 0.70 & 0.66 & 0.06 & \textbf{1.42} & \cellcolor{utilcol}0.66 & 0.72 & 0.64 & 0.76 & 0.71 & 0.02 & \textbf{1.85} & \cellcolor{utilcol}0.71 \\
 & SFT & 0.70 & 0.64 & 0.76 & 0.70 & 0.07 & 4.38 & \cellcolor{utilcol}0.69 & 0.70 & 0.64 & 0.76 & 0.70 & \textbf{0.05} & 4.22 & \cellcolor{utilcol}0.69 & 0.70 & 0.64 & 0.76 & 0.70 & 0.09 & 4.61 & \cellcolor{utilcol}0.69 \\
\rowcolor{ourrow}  & \ours{RO-PnR (Ours)} & \textbf{0.72} & \textbf{0.66} & \textbf{0.77} & \textbf{0.72} & \textbf{0.06} & 3.56 & \cellcolor{utilcol}\textbf{0.71} & \textbf{0.72} & \textbf{0.67} & \textbf{0.78} & \textbf{0.72} & 0.08 & 3.59 & \cellcolor{utilcol}\textbf{0.72} & 0.73 & \textbf{0.67} & \textbf{0.77} & \textbf{0.72} & 0.09 & 3.59 & \cellcolor{utilcol}\textbf{0.72} \\
\midrule
\multirow{6}{*}{\textbf{MisinfoCorrect}} & Single-Turn & 0.55 & 0.50 & 0.58 & 0.54 & 0.07 & 5.00 & \cellcolor{utilcol}0.53 & 0.55 & 0.49 & 0.59 & 0.54 & 0.05 & 5.00 & \cellcolor{utilcol}0.54 & 0.65 & 0.60 & 0.67 & 0.64 & \textbf{0.01} & 5.00 & \cellcolor{utilcol}0.63 \\
 & Fixed-Q & 0.70 & 0.63 & 0.72 & 0.69 & 0.07 & 5.00 & \cellcolor{utilcol}0.68 & 0.71 & 0.63 & 0.72 & 0.69 & \textbf{0.03} & 5.00 & \cellcolor{utilcol}0.68 & 0.71 & 0.64 & 0.71 & 0.69 & 0.03 & 5.00 & \cellcolor{utilcol}0.68 \\
 & Reactive & 0.71 & 0.64 & 0.74 & 0.70 & 0.06 & 4.98 & \cellcolor{utilcol}0.69 & 0.71 & 0.63 & 0.71 & 0.69 & 0.03 & 5.00 & \cellcolor{utilcol}0.68 & 0.70 & 0.63 & 0.70 & 0.68 & 0.03 & 3.96 & \cellcolor{utilcol}0.67 \\
 & Confidence & 0.61 & 0.54 & 0.64 & 0.60 & 0.08 & \textbf{1.10} & \cellcolor{utilcol}0.60 & 0.59 & 0.52 & 0.64 & 0.58 & 0.03 & \textbf{1.08} & \cellcolor{utilcol}0.58 & 0.63 & 0.55 & 0.69 & 0.62 & 0.02 & \textbf{1.35} & \cellcolor{utilcol}0.62 \\
 & SFT & 0.71 & 0.65 & 0.76 & 0.71 & \textbf{0.06} & 4.01 & \cellcolor{utilcol}0.70 & 0.71 & 0.65 & 0.76 & 0.70 & 0.07 & 3.91 & \cellcolor{utilcol}0.70 & 0.71 & 0.65 & 0.76 & 0.71 & 0.08 & 4.03 & \cellcolor{utilcol}0.70 \\
\rowcolor{ourrow}  & \ours{RO-PnR (Ours)} & \textbf{0.73} & \textbf{0.68} & \textbf{0.77} & \textbf{0.72} & 0.07 & 3.56 & \cellcolor{utilcol}\textbf{0.72} & \textbf{0.74} & \textbf{0.68} & \textbf{0.77} & \textbf{0.73} & 0.07 & 3.46 & \cellcolor{utilcol}\textbf{0.72} & \textbf{0.74} & \textbf{0.69} & \textbf{0.77} & \textbf{0.73} & 0.08 & 3.55 & \cellcolor{utilcol}\textbf{0.73} \\
\midrule
\multirow{6}{*}{\textbf{PUBHEALTH}} & Single-Turn & 0.63 & 0.58 & 0.65 & 0.62 & 0.24 & 5.00 & \cellcolor{utilcol}0.61 & 0.62 & 0.56 & 0.64 & 0.61 & 0.21 & 5.00 & \cellcolor{utilcol}0.60 & 0.66 & 0.61 & 0.68 & 0.65 & 0.09 & 5.00 & \cellcolor{utilcol}0.64 \\
 & Fixed-Q & 0.73 & 0.66 & 0.77 & 0.72 & 0.13 & 5.00 & \cellcolor{utilcol}0.71 & 0.73 & 0.67 & 0.75 & 0.72 & 0.12 & 5.00 & \cellcolor{utilcol}0.71 & 0.74 & 0.67 & 0.77 & 0.73 & \textbf{0.01} & 5.00 & \cellcolor{utilcol}0.72 \\
 & Reactive & 0.74 & 0.66 & 0.78 & 0.73 & 0.14 & 4.93 & \cellcolor{utilcol}0.72 & \textbf{0.75} & 0.68 & 0.75 & 0.73 & 0.12 & 5.00 & \cellcolor{utilcol}0.72 & 0.74 & 0.67 & 0.77 & 0.73 & \textbf{0.01} & 3.82 & \cellcolor{utilcol}0.72 \\
 & Confidence & 0.69 & 0.62 & 0.72 & 0.68 & 0.19 & \textbf{1.43} & \cellcolor{utilcol}0.68 & 0.71 & 0.64 & 0.73 & 0.69 & 0.14 & \textbf{1.63} & \cellcolor{utilcol}0.69 & \textbf{0.75} & 0.67 & \textbf{0.79} & \textbf{0.74} & 0.06 & \textbf{2.03} & \cellcolor{utilcol}\textbf{0.74} \\
 & SFT & 0.72 & 0.66 & 0.78 & 0.72 & \textbf{0.06} & 4.23 & \cellcolor{utilcol}0.72 & 0.73 & 0.67 & 0.78 & 0.72 & \textbf{0.05} & 4.24 & \cellcolor{utilcol}0.72 & 0.73 & 0.67 & 0.78 & 0.72 & 0.09 & 4.60 & \cellcolor{utilcol}0.72 \\
\rowcolor{ourrow}  & \ours{RO-PnR (Ours)} & \textbf{0.74} & \textbf{0.69} & \textbf{0.79} & \textbf{0.74} & 0.08 & 3.53 & \cellcolor{utilcol}\textbf{0.73} & 0.74 & \textbf{0.69} & \textbf{0.79} & \textbf{0.74} & 0.08 & 3.80 & \cellcolor{utilcol}\textbf{0.73} & 0.74 & \textbf{0.69} & 0.78 & 0.74 & 0.14 & 3.57 & \cellcolor{utilcol}0.73 \\
\bottomrule
\end{tabular}%
}
\caption{Average results across all health-literacy and belief-commitment configurations, for each method, dataset, and base model. \textbf{AA}, \textbf{PG}, \textbf{TA} are the Audience Alignment (AA), Personalized Grounding(PG), and Tailored Actionability (TA) scores (normalized to $[0,1]$); \textbf{Quality} is the overall quality of AA, PG and TA (normalized to $[0,1]$); \textbf{FER} is the factual error rate ($\downarrow$); \textbf{Turns} is the average number of dialogue turns ($\downarrow$); and \textbf{Utility} is the cost-adjusted Quality at $c{=}0.01$, our headline metric (normalized to $[0,1]$). 
}
\label{tab:main_results}
\end{table*}
\subsection{Dataset}
For fine-tuning and evaluation, we construct multi-turn datasets with a simulated user environment for health misinformation intervention, see user simulation in Section \ref{User Simulation} and dataset construction details in Appendix \ref{Appendix: Training Dataset Construction}. Our primary dataset is derived from CounterHealth (See Appendix \ref{Dataset Collection}), and is used for model fine-tuning. We additionally use two public health-misinformation datasets for evaluation, MisinfoCorrect\cite{he2023reinforcement} and PUBHEALTH~\cite{kotonya2020explainable}.

\subsection{User Simulation}
\label{User Simulation}
Due to the cost and difficulty of recruiting real users for large-scale multi-turn evaluation, we use an LLM-based user simulator to generate user actions during interaction. Specifically, we use GPT-4o-mini\footnote{\url{https://platform.openai.com/docs/models/gpt-4o-mini}}, which has been increasingly adopted as a proxy for simulating user behavior in interactive dialogue settings~\cite{kim2025can,laban2025llms,liu2026synthetic,lin2024diversedialogue}. In our setting, the simulator is prompted to emulate users with varying levels of health literacy and belief commitment, allowing us to evaluate how agents respond to different users. The user-simulation prompt is shown in Figure~\ref{fig:user-prompt}, with additional details provided in Appendix~\ref{User Simulation}.

To evaluate whether the LLM-based user simulator faithfully follows the assigned user profile, we conduct human evaluation along two dimensions: Persona Accuracy and Persona Consistency, referring to ~\cite{wang2025know}. Persona Accuracy measures whether the simulated user accurately reflects the assigned levels of health literacy and belief commitment. Persona Consistency measures whether the simulated user maintains the assigned health literacy and belief commitment across the dialogue. We instruct human annotators to rate each dimension on a 5-point scale. The detailed annotation rubric is provided in Table~\ref{tab:user-simulation-rubric}. The results are discussed in Section \ref{7: Human Validation}.
\subsection{Evaluation Metrics}
\label{Evaluation Metrics}

We evaluate system performance along three complementary dimensions: \emph{adaptation}, \emph{factual reliability}, and \emph{dialogue quality}, which together capture whether the system tailors the correction to the user, keeps it scientifically accurate, and reaches the goal efficiently. This goes beyond asking whether the final response is correct and also asks whether it is delivered in a way that is adaptive and effective.

\paragraph{Adaptation.} Effective health misinformation correction depends not only on \emph{what} is said but on \emph{how} it is framed for a specific user~\cite{krishnamurthy2026covid, nguyen2019tailored}. We assess this with three sub-metrics: \textit{Audience Alignment (AA)}~\cite{song2025speaking, cima2025contextualized}, which measures whether the correction is framed appropriately for the user's inferred health literacy and belief commitment; \textit{Personalized Grounding (PG)}~\cite{gabriel2024misinfoeval}, which measures whether the evidence and reasoning supporting the correction are clear and comprehensible to the user; and \textit{Tailored Actionability (TA)}~\cite{song2025speaking, ownby2026user}, which measures whether the agent provides safe, concrete next steps suited to the user when needed. All three are scored by an LLM judge (\textsc{mistral-large-2512\footnote{\url{https://docs.mistral.ai/models/model-cards/mistral-large-3-25-12}}}) given the dialogue history and the user's latent state. (See Figure~\ref{fig:twn-rubric} in the Appendix for detailed rubrics.) 

\paragraph{Factual reliability.} A correction that introduces new inaccuracies can undermine user trust and lead to harmful decisions~\cite{yue2024evidence}, so adaptation alone is not sufficient. We report the \textit{factual error rate}: the fraction of responses flagged as containing any factual error by an LLM judge (GPT-5-mini\footnote{\url{https://platform.openai.com/docs/models/gpt-5-mini}}) trained with updated knowledge and have access to web-based verification~\cite{he2023reinforcement}. Lower values indicate better performance. Prompt is in Figure~\ref{fig:fact-checking-judge-prompt} in the Appendix.

\paragraph{Dialogue quality.} In addition to adaptation metrics, we also measure the efficiency of the interaction. \textit{Turns}~\cite{wu2025collabllm} counts the number of communication turns, with fewer turns indicating a more efficient interaction. 
\textit{Utility}~\cite{dong2026value} captures the trade-off between response quality and interaction cost. Since the three adaptation dimensions are measured on the same scale and are jointly necessary for an adaptive response, we define overall ~\textit{Quality} as their mean, $Q = (S_{\text{AA}} + S_{\text{PG}} + S_{\text{TA}})/3$, and utility as $U = Q - c$, where $c$ is the cost incurred by probe actions.

\definecolor{cLcol}{HTML}{E8F4D9} 
\definecolor{cMcol}{HTML}{CFEAE0} 
\definecolor{cHcol}{HTML}{B8DCEB} 
\newcommand{\cL}[1]{\cellcolor{cLcol}#1}
\newcommand{\cM}[1]{\cellcolor{cMcol}#1}
\newcommand{\cH}[1]{\cellcolor{cHcol}#1}
\begin{table}[t]
\centering
\setlength{\tabcolsep}{1.5pt}
\renewcommand{\arraystretch}{1.05}
\scriptsize
\resizebox{\columnwidth}{!}{%
\begin{tabular}{@{}ll cccc cccc cccc@{}}
\toprule
 & & \multicolumn{4}{c}{\textbf{CounterHealth}} & \multicolumn{4}{c}{\textbf{Misinfo}} & \multicolumn{4}{c}{\textbf{Pubhealth}} \\
\cmidrule(lr){3-6} \cmidrule(lr){7-10} \cmidrule(lr){11-14}
\textbf{Method} & \textbf{Model} & F & I & C & Ov. & F & I & C & Q. & F & I & C & Ov. \\
\midrule
\multirow{3}{*}{Single-Turn}
 & Llama  & 0.51 & 0.60 & 0.57 & 0.56 & 0.49 & 0.57 & 0.55 & 0.53 & 0.57 & 0.64 & 0.62 & 0.61 \\
 & Qwen   & 0.50 & 0.59 & 0.59 & 0.56 & 0.48 & 0.56 & 0.56 & 0.54 & 0.54 & 0.63 & 0.62 & 0.60 \\
 & Gemma  & 0.54 & 0.66 & 0.67 & 0.62 & 0.56 & 0.67 & 0.67 & 0.63 & 0.56 & 0.67 & 0.68 & 0.64 \\
\midrule
\multirow{3}{*}{Fixed-Q}
 & Llama  & 0.66 & \cL{0.72} & \cL{0.72} & 0.70 & 0.62 & \cL{0.71} & \cL{0.70} & 0.68 & 0.67 & \cM{0.73} & \cM{0.73} & \cL{0.71} \\
 & Qwen   & 0.62 & \cL{0.71} & \cL{0.72} & 0.68 & 0.63 & \cL{0.71} & 0.70 & 0.68 & 0.67 & \cL{0.72} & \cM{0.73} & \cL{0.71} \\
 & Gemma  & 0.64 & \cL{0.72} & \cM{0.74} & 0.70 & 0.63 & \cL{0.70} & 0.70 & 0.68 & 0.67 & \cM{0.73} & \cH{0.75} & \cL{0.72} \\
\midrule
\multirow{3}{*}{Reactive}
 & Llama  & 0.65 & \cL{0.72} & \cL{0.72} & 0.69 & 0.64 & \cL{0.72} & \cL{0.71} & 0.69 & 0.67 & \cM{0.73} & \cM{0.74} & \cL{0.72} \\
 & Qwen   & 0.62 & \cL{0.71} & \cL{0.72} & 0.68 & 0.62 & \cL{0.71} & 0.70 & 0.68 & 0.67 & \cM{0.73} & \cM{0.74} & \cL{0.71} \\
 & Gemma  & 0.63 & \cL{0.72} & \cM{0.74} & 0.70 & 0.62 & 0.70 & 0.69 & 0.67 & 0.67 & \cM{0.73} & \cH{0.76} & \cL{0.72} \\
\midrule
\multirow{3}{*}{Confidence}
 & Llama  & 0.66 & \cL{0.72} & 0.70 & 0.69 & 0.60 & 0.63 & 0.56 & 0.60 & 0.65 & \cL{0.71} & 0.67 & 0.68 \\
 & Qwen   & 0.59 & \cL{0.71} & 0.68 & 0.66 & 0.57 & 0.62 & 0.56 & 0.58 & 0.64 & \cM{0.73} & \cL{0.70} & 0.69 \\
 & Gemma  & \cL{0.70} & \cL{0.72} & \cL{0.70} & \cL{0.71} & 0.66 & 0.64 & 0.56 & 0.62 & \cL{0.72} & \cM{0.74} & \cM{0.75} & \cM{0.74} \\
\midrule
\multirow{3}{*}{SFT}
 & Llama  & 0.64 & \cL{0.72} & \cL{0.72} & 0.69 & 0.64 & \cM{0.73} & \cM{0.73} & \cL{0.70} & 0.67 & \cM{0.74} & \cM{0.74} & \cL{0.71} \\
 & Qwen   & 0.64 & \cL{0.72} & \cL{0.72} & 0.69 & 0.63 & \cM{0.73} & \cL{0.72} & 0.70 & 0.66 & \cM{0.74} & \cM{0.74} & \cL{0.72} \\
 & Gemma  & 0.63 & \cL{0.72} & \cL{0.72} & 0.69 & 0.64 & \cM{0.73} & \cM{0.74} & \cL{0.70} & 0.66 & \cM{0.73} & \cM{0.75} & \cL{0.72} \\
\midrule
\multirow{3}{*}{RO-PnR}
 & Llama  & 0.65 & \cM{0.74} & \cM{0.74} & \cL{0.71} & 0.66 & \cM{0.75} & \cM{0.75} & \cL{0.72} & 0.68 & \cH{0.76} & \cH{0.77} & \cM{0.73} \\
 & Qwen   & 0.66 & \cM{0.75} & \cM{0.74} & \cL{0.72} & 0.66 & \cH{0.75} & \cH{0.75} & \cL{0.72} & 0.67 & \cH{0.76} & \cH{0.77} & \cM{0.73} \\
 & Gemma  & 0.65 & \cM{0.75} & \cH{0.75} & \cL{0.72} & 0.67 & \cH{0.75} & \cH{0.76} & \cM{0.73} & 0.67 & \cH{0.75} & \cH{0.77} & \cM{0.73} \\
\bottomrule
\end{tabular}%
} %
\caption{Utility by Health Literacy Level (normalized to $[0,1]$). Higher-value cells are highlighted with light tints: green ($\geq 0.70$), mint ($\geq 0.73$), and blue ($\geq 0.75$).
F, I, C, Ov., denotes Functional, Interactive, Critical, and all populations.
}
\label{tab:Health literacy Grouped Results}
\end{table}
\definecolor{cLcol}{HTML}{E8F4D9} 
\definecolor{cMcol}{HTML}{CFEAE0} 
\definecolor{cHcol}{HTML}{B8DCEB} 

\renewcommand{\cL}[1]{\cellcolor{cLcol}#1}
\renewcommand{\cM}[1]{\cellcolor{cMcol}#1}
\renewcommand{\cH}[1]{\cellcolor{cHcol}#1}

\begin{table}[t]
\centering

\setlength{\tabcolsep}{1.5pt}
\renewcommand{\arraystretch}{1.05}
\scriptsize
\resizebox{\columnwidth}{!}{%
\begin{tabular}{@{}ll cccc cccc cccc@{}}
\toprule
 & & \multicolumn{4}{c}{\textbf{CounterHealth}} & \multicolumn{4}{c}{\textbf{Misinfo}} & \multicolumn{4}{c}{\textbf{Pubhealth}} \\
\cmidrule(lr){3-6} \cmidrule(lr){7-10} \cmidrule(lr){11-14}
\textbf{Method} & \textbf{Model} & S & H & O & Ov. & S & H & O & Ov. & S & H & O & Ov. \\
\midrule
\multirow{3}{*}{Single-Turn}
 & Llama & 0.50 & 0.54 & 0.64 & 0.56 & 0.48 & 0.52 & 0.61 & 0.53 & 0.55 & 0.57 & \cL{0.71} & 0.61 \\
 & Qwen & 0.50 & 0.54 & 0.64 & 0.56 & 0.48 & 0.52 & 0.60 & 0.54 & 0.53 & 0.57 & 0.69 & 0.60 \\
 & Gemma & 0.57 & 0.62 & 0.69 & 0.62 & 0.58 & 0.62 & 0.69 & 0.63 & 0.58 & 0.62 & \cL{0.71} & 0.64 \\
\midrule
\multirow{3}{*}{Fixed-Q}
 & Llama & 0.63 & \cL{0.71} & \cH{0.75} & 0.70 & 0.60 & 0.69 & \cM{0.74} & 0.68 & 0.65 & \cL{0.72} & \cH{0.76} & \cL{0.71} \\
 & Qwen & 0.61 & \cL{0.71} & \cM{0.73} & 0.68 & 0.60 & \cL{0.70} & \cM{0.73} & 0.68 & 0.65 & \cL{0.72} & \cM{0.75} & \cL{0.71} \\
 & Gemma & 0.61 & \cL{0.73} & \cH{0.76} & 0.70 & 0.57 & \cL{0.71} & \cH{0.75} & 0.68 & 0.65 & \cM{0.74} & \cH{0.76} & \cL{0.72} \\
\midrule
\multirow{3}{*}{Reactive}
 & Llama & 0.62 & \cL{0.71} & \cH{0.76} & 0.69 & 0.62 & 0.70 & \cH{0.75} & 0.69 & 0.66 & \cM{0.73} & \cH{0.76} & \cL{0.72} \\
 & Qwen & 0.61 & 0.70 & \cM{0.74} & 0.68 & 0.60 & 0.69 & \cM{0.74} & 0.68 & 0.65 & \cM{0.73} & \cH{0.76} & \cL{0.72} \\
 & Gemma & 0.62 & \cL{0.72} & \cH{0.76} & 0.70 & 0.57 & 0.69 & \cM{0.75} & 0.67 & 0.65 & \cM{0.74} & \cH{0.77} & \cL{0.72} \\
\midrule
\multirow{3}{*}{Confidence}
 & Llama & 0.65 & 0.70 & \cM{0.73} & 0.69 & 0.58 & 0.59 & 0.62 & 0.60 & 0.65 & 0.68 & \cL{0.71} & 0.68 \\
 & Qwen & 0.63 & 0.66 & 0.69 & 0.66 & 0.57 & 0.57 & 0.60 & 0.58 & 0.64 & 0.70 & \cM{0.73} & 0.69 \\
 & Gemma & 0.67 & \cL{0.71} & \cM{0.73} & \cL{0.71} & 0.61 & 0.62 & 0.64 & 0.62 & 0.69 & \cH{0.75} & \cH{0.77} & \cM{0.74} \\
\midrule
\multirow{3}{*}{SFT}
 & Llama & 0.62 & \cL{0.71} & \cM{0.75} & 0.69 & 0.64 & \cL{0.72} & \cH{0.76} & \cL{0.70} & 0.65 & \cM{0.73} & \cH{0.76} & \cL{0.71} \\
 & Qwen & 0.63 & \cL{0.71} & \cM{0.75} & 0.69 & 0.63 & \cL{0.71} & \cM{0.75} & 0.70 & 0.66 & \cM{0.73} & \cH{0.76} & \cL{0.72} \\
 & Gemma & 0.62 & \cL{0.71} & \cM{0.75} & 0.69 & 0.64 & \cL{0.71} & \cM{0.75} & 0.70 & 0.66 & \cM{0.73} & \cH{0.76} & \cL{0.72} \\
\midrule
\multirow{3}{*}{RO-PnR}
 & Llama & 0.65 & \cL{0.72} & \cH{0.76} & \cL{0.71} & 0.66 & \cL{0.72} & \cH{0.77} & \cL{0.72} & 0.69 & \cM{0.74} & \cH{0.77} & \cM{0.73} \\
 & Qwen & 0.66 & \cL{0.72} & \cH{0.76} & \cL{0.72} & 0.67 & \cM{0.73} & \cH{0.77} & \cL{0.72} & 0.69 & \cM{0.74} & \cH{0.77} & \cM{0.73} \\
 & Gemma & 0.67 & \cL{0.72} & \cH{0.76} & \cL{0.72} & 0.69 & \cM{0.73} & \cH{0.77} & \cM{0.73} & 0.69 & \cM{0.73} & \cH{0.77} & \cM{0.73} \\
\bottomrule
\end{tabular}%
} %
\caption{Utility by Belief Commitment. Higher-value cells are highlighted with light tints: green ($\geq 0.70$), mint ($\geq 0.73$), and blue ($\geq 0.75$); 
lower values are left uncolored. 
S, H, O, Ov., denotes Strong, Hesitant, Open, and all populations.
}
\label{tab:Belief Commitment Grouped Results}

\end{table}

\subsection{Baselines}
We compare our RO-PnR framework against four alternative decision strategies:
(a) \textit{Single-Turn Direct Response (Single-Turn)}~\cite{song2025speaking}: The model generates the correction immediately based only on the current turn, without asking clarification questions. (b) \textit{Fixed-Question Probing (Fixed-Q)}~\cite{fu2025first}: The agent asks a predetermined number of clarification questions before producing the final counterspeech response, regardless of the user’s feedback.
(c) \textit{Reactive Clarifier (Reactive)}~\cite{zhang2025clarify}: The agent adaptively chooses between asking and answering based on user feedback, without considering the user's knowledge background or stance toward the claim.
(d) \textit{Confidence-Gated Clarifier (Confidence)}~\cite{NEURIPS2024_32b80425}: The agent decides whether to ask or answer based on its confidence in a generic inferred user state, not grounded in the health literacy and belief commitment dimensions central to misinformation correction.
(e) \textit{Supervised Clarifier (SFT)}: A method that learns better to ask and answer via supervised fine-tuning on a higher-quality trajectory, without the turn-level decision reward or GRPO optimization. 

\subsection{Implementation Details}

We evaluate RO-PnR on three open-source 8B models: \textsc{Llama-3.1-8B-Instruct}\footnote{\url{https://huggingface.co/meta-llama/Llama-3.1-8B-Instruct}}, \textsc{Gemma-4-E4B-it}\footnote{\url{https://huggingface.co/google/gemma-4-E4B-it}}, and \textsc{Qwen3-8B}\footnote{\url{https://huggingface.co/Qwen/Qwen3-8B}}. 
Fine-tuning consists of two stages: offline supervised fine-tuning (SFT) and offline GRPO~\cite{shao2024deepseekmath}.
We first perform SFT on base models using full-profile dialogue examples with LoRA adapters~\cite{hu2022lora}. We then initialize GRPO from the resulting SFT adapter and continue training on turn-level preference examples constructed from all nine user-profile combinations. Training configurations are provided in Table \ref{tab:training_config}.

\section{Results}

\paragraph{Main Results.}
RO-PnR achieves the best overall utility while using substantially fewer interaction turns than always-probe baselines (Table~\ref{tab:main_results}). 
RO-PnR obtains the highest utility on nearly every (dataset, model) combination, reaching 0.71--0.73 in utility versus 0.68-0.72 for the strongest baseline (SFT), with consistent gains across all three adaptation dimensions (AA, PG, TA). The baselines reveal a clear trade-off. Single-Turn skips probing entirely and scores lowest (0.53-0.64). Fixed-Q and Reactive almost probe every turn ($\geq 4.9$ on average) and reach competitive adaptation, but at a high interaction cost. Confidence rarely (1.1-2.0 turns) and suffers the largest quality drop, especially on MisinfoCorrect, where utility falls to 0.58-0.62. RO-PnR occupies the productive middle, using roughly 3.5 turns, about 30\% fewer than the always-probe baselines, while delivering the highest utility. Factual error rates remain low across methods (0.06-0.14 for RO-PnR), indicating that adaptation gains do not compromise factual reliability.

\paragraph{Performance Across User Profiles.}
RO-PnR's utility gains hold across user types, with the largest improvements on the hardest user slices. Tables~\ref{tab:Health literacy Grouped Results} and~\ref{tab:Belief Commitment Grouped Results} break utility down by health literacy and belief commitment. RO-PnR achieves the highest overall utility in every (dataset, model) cell, and its advantage is most pronounced on the most challenging slices. On Functional users, where adaptive framing matters most, RO-PnR reaches 0.65-0.69 versus 0.62-0.67 for SFT. On Strong-believer users, the gap widens further: RO-PnR obtains 0.65--0.69 versus 0.6-0.66 for SFT and only 0.57-0.61 for Confidence, indicating that indiscriminate probing fails to overcome resistance while confidence-only probing fails to gather sufficient context. As literacy rises and belief commitment weakens, all methods improve, and the gap narrows, e.g., on Open users, RO-PnR reaches 0.76-0.77 while baselines cluster in the 0.73-0.76 range, but RO-PnR retains the top spot throughout. These breakdowns confirm that the utility gains in the main results are not driven by a single easy slice but hold across both the comprehension and the openness dimensions of user heterogeneity.

\section{Ablation Study}
\label{Ablation Study}
\begin{table}[t]
\centering
\small
\setlength{\tabcolsep}{5pt}
\renewcommand{\arraystretch}{1.15}
\resizebox{\linewidth}{!}{
\begin{tabular}{l ccc c cc c}
\toprule
\textbf{Variant} & \textbf{AA} $\uparrow$ & \textbf{PG} $\uparrow$ & \textbf{TA} $\uparrow$ & \textbf{Quality} $\uparrow$ & \textbf{FER} $\downarrow$ & \textbf{Turns} $\downarrow$ & \textbf{Utility} $\uparrow$ \\
\midrule
\textbf{RO-PnR (full)} & \textbf{0.72} & \textbf{0.66} & \textbf{0.77} & \textbf{0.72} & \textbf{0.06} & 3.56 & \textbf{0.71} \\
\midrule
\multicolumn{8}{l}{\textit{(a) Reward-horizon ablation}} \\
\quad immediate next-gain only & 0.62 & 0.54 & 0.65 & 0.60 & 0.12 & 2.00 & 0.60 \\
\midrule
\multicolumn{8}{l}{\textit{(b) Probe-cost ($c$) sweep}} \\
\quad probe cost $= 0.00$ & 0.50 & 0.46 & 0.57 & 0.51 & 0.11 & 5.83 & 0.51 \\
\quad probe cost $= 0.03$ & 0.71 & 0.65 & 0.77 & 0.71 & 0.08 & 3.00 & 0.70 \\
\quad probe cost $= 0.05$ & 0.72 & 0.66 & 0.77 & 0.71 & 0.09 & 3.00 & 0.69 \\
\bottomrule
\end{tabular}}
\caption{Ablation study results, averaged over all (health literacy $\times$ belief commitment) user profiles. \textbf{RO-PnR (full)} is our complete method with forward-looking long-horizon reward, probe cost $c = 0.01$, and cross-threshold bonus. (a) replaces the forward-looking reward with the immediate next-turn gain; (b) sweeps the per-turn probe cost $c$.
}
\label{tab:ablation}
\end{table}

To investigate how different components contribute to RO-PnR performance, we conduct two ablations on the GRPO reward (Table~\ref{tab:ablation}). \textbf{(a) Reward-horizon ablation} replaces the forward-looking reward with a one-step reward based only on the next user response, testing whether short-horizon credit assignment suffices. \textbf{(b) Probe-cost sweep} varies the per-question cost $c$ to probe the trade-off between information gathering and early commitment. 

Table~\ref{tab:ablation} confirms that all components contribute to performance. The reward-horizon ablation degrades utility from 0.71 to 0.60, showing that a probe's value often materializes several turns later and cannot be captured by a one-step reward. The probe-cost sweep shows that $c = 0$ leads to over-probing (5.83 turns) and a quality collapse to 0.51, while moderate costs ($c = 0.03$-$0.05$) recover near-best performance at ${\sim}3$ turns, confirming that the policy is stable across a reasonable range of $c$. 

\section{Human Validation}
\label{7: Human Validation}
While LLMs are known to reliably simulate humans and act as judges at low cost~\cite{thakur2025judging, bavaresco2025llms, huang2025empirical}, we run a small-scale human evaluation along three axes: (1) simulator reliability via persona accuracy and consistency; (2) human-judge agreement on adaptation ratings against \textsc{Mistral-Large-2512} (rubric in Figure~\ref{fig:twn-rubric}); and (3) a pairwise preference study comparing \textsc{RO-PnR} against the baseline (Due to the annotation cost, we include only the strong baseline method: Fixed-Q.) with six users spanning distinct health literacy and belief commitment profiles. Details in Appendix~\ref{Human Validation}.

\paragraph{LLM Simulator Evaluation.}
Annotators show substantial agreement on both axes (Table~\ref{tab:persona_agreement}): mean pairwise agreement is 0.83 for persona accuracy and 0.85 for consistency, with multi-rater Scott's $\pi$ of 0.70 and 0.68. Using the median of the three annotators' ratings as the final score (Table~\ref{tab:user_simulation_rating_distribution}), persona accuracy averages 3.87, with 76\% of samples scoring 4 or 5, indicating that simulated users generally reflect the assigned health literacy and belief commitment, though many are judged as largely rather than perfectly aligned. Persona consistency is stronger, averaging 4.64 with 97\% of samples at 4 or above, showing that once a persona is established, the simulator maintains it across turns. Details are in Appendix~\ref{Appendix: LLM Simulator Evaluation}.

\paragraph{Human-judge agreement.}
Human annotators show moderate-to-strong internal consistency (Scott's $\pi$ = 0.63--0.78 across dimensions; Table~\ref{tab:human_internal_agreement for AA_PG_TA}). Using the average human rating as consensus, the LLM judge aligns well with humans on \textit{Quality}: Scott's $\pi$ = 0.55, 73\% exact agreement, MAE = 0.23, and 87\% of samples within $\pm 0.5$ points (Table~\ref{tab:Human_LLM_agreement}), with Pearson = 0.57 and Spearman = 0.72 (Table~\ref{tab:judge_human_correlation}). This indicates that the LLM judge is a reliable proxy for human evaluation at scale.

\paragraph{Human pairwise preference.}
As shown in Figure~\ref{fig:PnR_vs_FixedQ}, RO-PnR is strongly preferred overall (79.70\% vs.\ 20.30\% for \textit{Fixed-Q}), primarily because its responses are perceived as more explanatory, credible, and actionable. The exception is User~2 (functional literacy, strong belief commitment), who favors \textit{Fixed-Q} for being easier to process, less confrontational, and less reliant on institutional authority, consistent with prior findings that users distrustful of official health sources resist authority-heavy framings even when the evidence is richer~\cite{jamison2019you}. This highlights a trade-off: while \textsc{RO-PnR} yields higher-quality corrections in aggregate, users with strong prior commitments may benefit from softer, less institution-centered framing to reduce resistance.

\section{Conclusion}
In this paper, we introduced RO-PnR, a decision framework for multi-turn health misinformation intervention that learns when to ask clarifying questions and when to correct. By modeling hidden user differences in health literacy and belief commitment, RO-PnR adapts its probing behavior to the needs of each dialogue rather than relying on fixed or excessive questioning. Experiments show that RO-PnR achieves stronger cost-adjusted utility while using fewer interaction turns than always-probe baselines. These results suggest that effective misinformation correction requires not only accurate evidence, but also careful timing: asking when clarification is useful, and answering when enough context has been gathered. Future work should extend this framework with real-user interactions, richer models of user burden, and joint optimization of both when and how to probe.

\section*{Limitations}
\paragraph{Simulated User Interaction.}
Because collecting large-scale real-user interactions is costly, we rely on simulated users as a proxy for human behavior. While prior studies suggest that LLM-based simulation can provide a useful approximation~\cite{wang2025know, sekulic2024reliable, bougie2025simuser}, and we make efforts to improve its reliability, it may still fall short of capturing the full nuance and variability of real users. Despite this limitation, our simulation framework offers a practical testbed for studying how health-misinformation-related user characteristics affect multi-turn counterspeech. Future work should incorporate real-user data and further refine both the simulator and the policy under more realistic interaction settings.

\paragraph{Probe Action Constraints.}
Our study focuses primarily on the decision of \emph{when} to probe and \emph{when} to stop, emphasizing the probe-and-respond policy rather than the content of the probe itself. As a result, the question of \emph{how} to probe, namely, what clarification question is most appropriate at a particular stage of the dialogue, remains open. Future work could address this limitation by jointly optimizing probe timing and probe content so that the agent can ask more adaptive and informative questions throughout the interaction.

\paragraph{Simplified Cost Modeling.}
We model communication cost following \citet{dong2026value} to account for the burden imposed by additional clarification turns. However, this formulation is still coarse-grained, assigning a fixed cost at the turn level rather than capturing the more nuanced cognitive load experienced by users. Although our study provides insight into how agents behave when communication cost is incorporated into the decision process, future work might explore more refined ways of measuring user burden and integrating it into the policy for more cognitively aware interaction.

\bibliography{custom}

\appendix
\clearpage
\section{User Modeling}
\label{Appendix: User Modeling Definition}

\subsection{Health literacy}
\label{Appendix: Health literacy}
\textbf{Functional health literacy} means basic reading/writing and understanding skills sufficient to function effectively in everyday health situations, such as understanding health information and service instructions~\cite{nutbeam2000health}.

\noindent\textbf{Interactive health literacy} is more advanced personal, communicative, and social skills that enable independent action on health knowledge~\cite{nutbeam2000health}.

\noindent\textbf{Critical health literacy}
This is the most advanced level. ~\citet{nutbeam2000health} defines it in terms of advanced cognitive and social skills used to critically analyze information and act on broader determinants of health at individual and community levels.

\begin{figure*}
    \centering
    \includegraphics[width=0.88\linewidth]{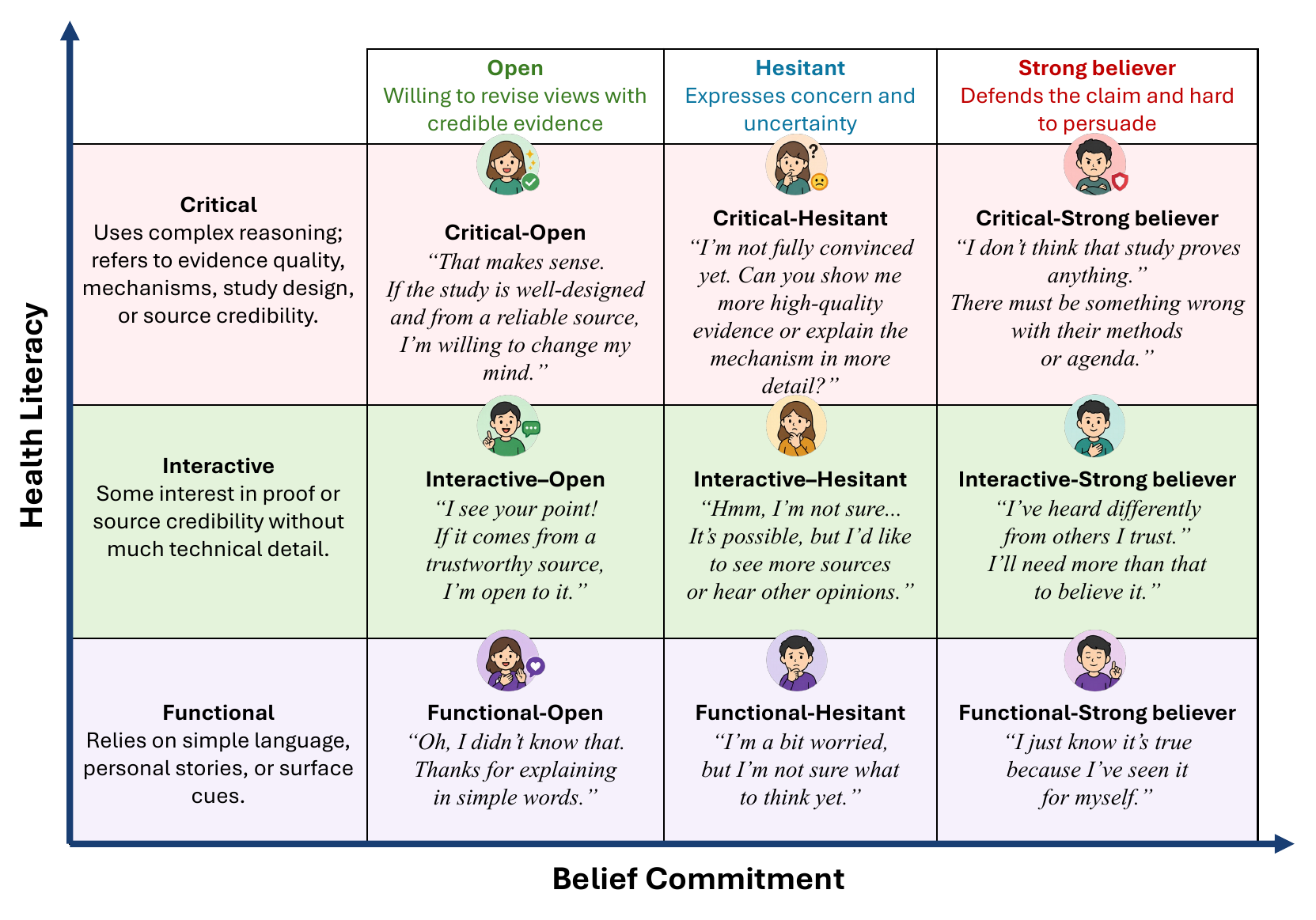}
    \caption{\textbf{3$\times$3 latent user profile grid used in our simulation environment.} Users are characterized along two dimensions: \emph{health literacy} (\textsc{Functional}, \textsc{Interactive}, \textsc{Critical}) and \emph{belief commitment} (\textsc{Open}, \textsc{Hesitant}, \textsc{Strong believer}). Each cell defines one latent user type and includes an illustrative response style showing how the user may express concern, interpret evidence, and respond to counterspeech. This profile grid is used to simulate heterogeneous user behavior during multi-turn interaction.}
    \label{fig:9_profile}
\end{figure*}

\begin{table*}[t]
\centering
\small
\resizebox{0.96\linewidth}{!}{
\begin{tabular}{l c c c c c p{6cm}}
\toprule
Stage & LR & Epochs & Batch & Grad. accum. & Max len. & Other settings \\
\midrule
SFT & $5\times10^{-5}$ & 4 & 1 & 8 & 2048 & LoRA $r=16$, $\alpha=32$, dropout $0.05$; bf16; 100 warmup steps; seed 42 \\
Offline GRPO & $5\times10^{-6}$ & 2 & 1 & 8 & 2048 &$\beta_{\mathrm{KL}}=0.05$; SFT CE weight $0.2$; warmup ratio $0.03$; max grad norm $1.0$; seed 42 \\
\bottomrule
\end{tabular}}
\caption{Hyperparameters for SFT and GRPO configuration}
\label{tab:training_config}
\end{table*}

\subsection{Belief Commitment}
\label{Belief Commitment}

\noindent\textbf{Strong belief commitment} can be understood as a state in which individuals accept misinformation as true, allow it to continue shaping subsequent attitudes and judgments~\cite{newman2022misinformation}, and show difficulty revising that belief even after corrective information is provided~\cite{wittenberg2020misinformation}. In stronger cases, correction may even increase belief in the original misconception, indicating an especially resistant form of commitment~\cite{swire2022backfire}.

\noindent\textbf{Hesitant belief commitment} is a state in which a person gives a misinformation claim partial acceptance, but does so with noticeable uncertainty or limited confidence~\cite{guigon2026rethinking, schulz2025understanding}, so the belief is not fully consolidated and remains comparatively more open to revision than a strong misinformation belief~\cite{kemp2024role}. 

\noindent\textbf{Open belief commitment} refers to a revisable form of belief commitment in which individuals may initially accept a claim, but remain willing to consider alternative evidence, evaluate source credibility, and update their beliefs when credible corrective information becomes available~\cite{jongman2023intellectual,metz2023building}. This form of commitment is better understood as an orientation toward revisability and evidence-based updating, rather than simply as weak belief~\cite{sanna2025belief}.

\subsection{User Simulation}
\label{User Simulation}

\begin{table*}[t]
\centering
\small
\setlength{\tabcolsep}{4pt}
\renewcommand{\arraystretch}{1.15}
\begin{tabular}{p{0.07\textwidth} p{0.43\textwidth} p{0.43\textwidth}}
\toprule
\textbf{Score} & \textbf{Persona Accuracy} & \textbf{Persona Consistency} \\
\midrule
1 &
The simulated user clearly does not match the assigned health literacy or belief commitment. &
The user frequently contradicts the assigned profile, with abrupt and unjustified shifts in knowledge level or belief strength. \\

2 &
The simulated user weakly matches the assigned profile. One dimension may be partially reflected, but the other is missing or incorrect. &
The user shows unstable behavior, with multiple unexplained shifts in health literacy or belief commitment. \\

3 &
The simulated user partially matches the assigned profile. Both dimensions are somewhat recognizable, but the dialogue is vague, generic, or only weakly aligned. &
The user is mostly stable, but there are noticeable inconsistencies or minor contradictions across turns. \\

4 &
The simulated user largely matches the assigned health literacy and belief commitment, with only minor imperfections. &
The user maintains the assigned profile across most turns, and any change in belief strength is mostly justified by the conversation. \\

5 &
The simulated user strongly matches the assigned profile. Health literacy and belief commitment are both clearly and accurately expressed. &
The user consistently maintains the assigned health literacy and belief commitment throughout the dialogue, with any change being natural and well-supported by the interaction. \\
\bottomrule
\end{tabular}
\caption{Human evaluation rubric for assessing user simulation quality.}
\label{tab:user-simulation-rubric}
\end{table*}

We simulate user replies with a profile-conditioned LLM environment. For the main results, each dialogue is paired with a fixed latent user profile
\[
u = (\text{health\_literacy}, \text{belief\_commitment}),
\]
and the same profile is kept fixed throughout the entire interaction. We evaluate all posts under the full $3 \times 3$ grid of user types: health literacy in \{\textit{Functional}, \textit{Interactive}, \textit{Critical}\} and belief commitment in \{\textit{Open}, \textit{Hesitant}, \textit{Strong}\}.

The profile dimensions are defined behaviorally rather than demographically. Referring to the definition of \citet{nutbeam2000health}, we specify that \textit{Health literacy} controls how complex the user’s reasoning sounds: Functional-literacy users prefer simple language and rely more on stories or surface cues; Interactive-literacy users show some interest in proof or source credibility without much technical detail; Critical-literacy users are more likely to refer to evidence quality, mechanisms, study design, or source credibility. Additionally, referring to section \ref{Belief Commitment}, we detail that \textit{Belief commitment} controls how resistant the user is to correction: strong believers defend the claim and are hard to persuade, hesitant users express concern and uncertainty at the same time, and open users are comparatively willing to revise their view if shown credible evidence.

At each turn, the simulator receives the fixed latent state, the dialogue history, and the assistant's latest question, and generates one short reply. In our main setting, the simulator is implemented with \texttt{gpt-4o-mini} at temperature $0.3$. The prompt explicitly conditions on the latent user state but instructs the model not to reveal it directly. Its structure is in Figure \ref{fig:user-prompt}.



\begin{table}[t]
\centering
\small
\setlength{\tabcolsep}{8pt}
\renewcommand{\arraystretch}{1.2}
\resizebox{\linewidth}{!}{
\begin{tabular}{@{}lcc@{}}
\toprule
\textbf{Metric} & \textbf{Persona Accuracy} & \textbf{Persona Consistency} \\
\midrule
Mean Pairwise Agreement     & 0.83 & 0.85 \\
Mean Pairwise Scott's $\pi$ & 0.70 & 0.68 \\
\addlinespace[2pt]
All-Three Agreement         & 0.74 & 0.78 \\
Multi-Rater Scott's $\pi$   & 0.70 & 0.68 \\
\bottomrule
\end{tabular}}
\caption{Inter-annotator agreement among three evaluators on \textit{persona accuracy} and \textit{persona consistency}. Both pairwise and multi-rater statistics indicate substantial agreement.}
\label{tab:persona_agreement}
\end{table}




\begin{table}[t]
\centering
\small
\setlength{\tabcolsep}{5pt}
\renewcommand{\arraystretch}{1.12}
\resizebox{\linewidth}{!}{
\begin{tabular}{lcccccc}
\toprule
\textbf{Metric} & \textbf{Score 1} & \textbf{Score 2} & \textbf{Score 3} & \textbf{Score 4} & \textbf{Score 5} & \textbf{Mean} \\
\midrule
Persona Accuracy    & 0\% & 3\% & 21\% & 62\% & 14\% & 3.87 \\
Persona Consistency & 0\% & 0\% & 3\%  & 30\% & 67\% & 4.64 \\
\bottomrule
\end{tabular}}
\caption{Distribution of human ratings for persona accuracy and persona consistency. }
\label{tab:user_simulation_rating_distribution}
\end{table}

\begin{table}[t]
\centering
\small
\setlength{\tabcolsep}{6pt}
\renewcommand{\arraystretch}{1.12}
\resizebox{\linewidth}{!}{
\begin{tabular}{lccc}
\toprule
\textbf{Dimension} & \textbf{Scott's $\pi$} & \textbf{Mean Pairwise Agreement} & \textbf{All-three Exact Agreement} \\
\midrule
Audience Alignment                & 0.76 & 84.70\% & 78.00\% \\
Personalized Grounding                & 0.63 & 76.00\% & 65.00\% \\
Tailored Actionability                & 0.77 & 88.70\% & 83.00\% \\
Quality & 0.78 & 86.70\% & 81.00\% \\
\bottomrule
\end{tabular}}
\caption{Human inter-annotator agreement under 0.5-point discretization.}
\label{tab:human_internal_agreement for AA_PG_TA}
\end{table}

\begin{table}[t]
\centering
\small
\setlength{\tabcolsep}{5pt}
\renewcommand{\arraystretch}{1.12}
\resizebox{\linewidth}{!}{
\begin{tabular}{lcccccc}
\toprule
\textbf{Dim} & \textbf{Scott's $\pi$} & \textbf{Percent Agreement} & \textbf{MAE} & \textbf{Bias} & \textbf{Within $\pm 0.5$} & \textbf{Within $\pm 1.0$} \\
\midrule
Audience Alignment & 0.51 & 69.00\% & 0.29 & -0.14 & 86.00\% & 94.00\% \\
Personalized Grounding  & 0.31 & 51.00\% & 0.28 & -0.14 & 85.00\% & 96.00\% \\
Tailored Actionability  & 0.48 & 72.00\% & 0.14 & -0.07 & 95.00\% & 99.00\% \\
Quality & 0.55 & 73.00\% & 0.23 & -0.11 & 87.00\% & 97.00\% \\
\bottomrule
\end{tabular}}
\caption{Agreement between the human consensus and the  LLM judge across rubric dimensions and the aggregated final score. Bias is computed as consensus minus LLM score.}
\label{tab:Human_LLM_agreement}
\end{table}

\textbf{The prompt further specifies style and behavioral constraints.} It requires casual, brief replies in 1--2 sentences, discourages technical medical explanation unless it would arise naturally for that user type, and asks the model to make both health literacy and belief commitment observable through wording, evidence preferences, and openness to update. Separate behavior rules are injected for each profile. For example, functional-literacy users are told to use short everyday wording and focus on personal or practical concerns, whereas critical-literacy users are told to show stronger awareness of evidence quality and source credibility. Strong believers are instructed to sound firm and skeptical of correction, hesitant users to sound torn and uncertain, and open users to sound receptive and non-defensive.

\textbf{To reduce role drift, we combine several safeguards.} First, the system prompt explicitly says: the model is simulating a social media user, must stay strictly in the \emph{user} role, and must not act as an expert, educator, or assistant. Second, the user prompt reinforces this with strict stylistic constraints: short replies, no belief-state summary, no multiple questions, and no unnecessary factual exposition. Third, the model is required to return JSON only of the form \texttt{\{"reply": "..."\}}, which reduces conversational spillover and makes parsing more robust. Finally, because the latent state is fixed across turns, the simulator is encouraged to remain behaviorally consistent even as the dialogue evolves.

An important setting is that the assistant does not observe these profile labels explicitly. The user profile is hidden from the policy and only affects the environment-side reply generation. The assistant must infer how to adapt from the user’s language and reactions, rather than from direct access to the profile metadata.

\section{Training Dataset Construction}
\label{Appendix: Training Dataset Construction}

We construct training data to match the reward design at the level of \emph{turn-level decisions}, rather than treating each dialogue as a single terminal example. The starting point is a collection of scored multi-turn rollouts generated for each misinformation post under each of the nine user-profile conditions. For a fixed post--profile pair, we collect multiple stochastic trajectories, so the rollout set can be viewed as a set of \emph{empirical future samples}: some trajectories stop early and respond immediately, while others continue probing and reveal the value of obtaining additional user information.

From each trajectory, we extract prefix-level decision states $h_t$ from the first four turns. Each state is converted into a structured assistant target consisting of a latent belief estimate, an action token (\textsc{Probe} or \textsc{Respond}), and the associated natural-language content. Crucially, the rollout trace provides both the value of responding under the \emph{current} information state and the downstream outcomes that become reachable if the dialogue continues. We therefore build two kinds of supervision. A \textsc{Respond} record uses the candidate response available at state $h_t$ and is assigned the immediate stopping value $R_{\mathrm{now}}$. A \textsc{Probe} record uses the clarification question asked at that state and is assigned a forward-looking continuation value based on the best downstream outcome reachable after probing:
This framing makes each training example a local decision problem: should the policy stop now, or should it invest one more turn to access a better future response?

To ensure meaningful comparison, we organize records by source post, turn index, and user-profile condition. This groups together alternative trajectories that correspond to closely matched decision contexts, allowing \textsc{Probe} and \textsc{Respond} to be compared locally rather than across unrelated dialogues. We discard groups that are too small or exhibit little reward variation, and then normalize rewards within each group to obtain relative advantages for policy optimization. The resulting dataset is therefore not simply a set of high-quality responses; it is a structured decision dataset in which \textsc{Respond} is supervised by the current stopping value, while \textsc{Probe} is supervised by future-sampled continuation value. This organization makes the subsequent optimization directly aligned with the intended probe-and-respond behavior.

For SFT, we select the higher \textit{Quality} and factually clean trajectory for each post/profile group. The selected trajectories are converted into step-level supervision examples: given the visible conversation prefix, the model learns to output a structured assistant action, either probe with a clarification question or respond with a final response.

For GRPO, we use scored rollouts from all nine user-profile combinations. Each rollout is decomposed into turn-level decision examples. At each turn, candidate assistant actions are grouped by post, turn index, and profile, and assigned rewards using future reward. Groups with insufficient reward variation or without both probe and respond alternatives are filtered out. The remaining grouped examples provide normalized advantages for offline GRPO, initialized from the SFT adapter.

\section{CounterHealth Dataset Collection}
\label{Dataset Collection}
We collected Reddit posts and comments on COVID-19, influenza, and HIV via the PRAW API\footnote{\url{https://praw.readthedocs.io/}}, using health-related keywords (e.g., ``vaccines,'' ``COVID-19,'' ``alternative medicine'') to retrieve 4,968 posts and 17,223 comments from high-engagement subreddits. To identify misinformation, five trained annotators from information science backgrounds labeled a 1,000-post sample under a shared annotation guideline, yielding 330 confirmed health-misinformation posts. We then fine-tuned a RoBERTa-large classifier on these annotations (F1 $=0.76$) and applied it to the remaining posts, surfacing 831 additional candidates. A final filtering pass using GPT-5-mini\footnote{\url{https://platform.openai.com/docs/models/gpt-5-mini}} with web-search-assisted human review produced 769 high-quality health-misinformation posts. On a 100-post validation sample, three annotators showed substantial agreement with the final labels (mean pairwise agreement $= 87.4\%$, Cohen's $\kappa \geq 0.67$), confirming the reliability of the filtering pipeline.

\section{Human Validation}
\label{Human Validation}

\begin{table}[t]
\centering
\small
\setlength{\tabcolsep}{6pt}
\renewcommand{\arraystretch}{1.12}
\resizebox{\linewidth}{!}{
\begin{tabular}{lcccc}
\toprule
\textbf{Dimension} & \textbf{Pearson} & \textbf{Spearman} & \textbf{LLM Mean} & \textbf{Human Mean} \\
\midrule
Audience Alignment       & 0.50 & 0.66 & 3.76 & 3.62 \\
Personalized Grounding       & 0.51 & 0.67 & 3.50 & 3.37 \\
Tailored Actionability      & 0.72 & 0.81 & 3.93 & 3.86 \\
Quality & 0.57 & 0.72 & 3.73 & 3.62 \\
\bottomrule
\end{tabular}}
\caption{Correlation between the LLM judge and human consensus under 0.5-point discretization.}
\label{tab:judge_human_correlation}
\end{table}

\begin{table*}[htbp]
\centering
\setlength{\tabcolsep}{6pt}
\renewcommand{\arraystretch}{1.2}
\resizebox{\linewidth}{!}{
\begin{tabular}{lll}
\hline
\textbf{Item} & \textbf{Question} & \textbf{Rating} \\
\hline
B & How accurate do you think this claim is? 
  & 0 = Definitely false / 10 = Definitely true \\
C & How certain are you about your answer above? 
  & 0 = Not at all / 10 = Completely \\
O & If you saw credible evidence against your view on this claim, how willing would you be to change your mind? 
  & 0 = Not at all willing / 10 = Completely willing \\
\hline
\end{tabular}}
\caption{Belief Commitment Determinants. B denotes belief, C denotes certainty, and O denotes openness to change.}
\label{tab:belief_commitment_items}
\end{table*}

\begin{table}[htbp]
\centering
\setlength{\tabcolsep}{6pt}
\renewcommand{\arraystretch}{1.3}
\resizebox{\linewidth}{!}{
\begin{tabular}{lcccl}
\hline
\textbf{User} & \textbf{Total FCCHL Mean} & \textbf{HL Category} & \textbf{Overall \(BC_{\text{total}}\)} & \textbf{BC Category} \\
\hline
User 1 & 2.07 & Communicative & 2.03 & Hesitant \\
User 2 & 1.93 & Functional & 6.83 & Strong \\
User 3 & 2.71 & Communicative & 1.20 & Open \\
User 4 & 3.00 & Critical & 1.20 & Open \\
User 5 & 4.00 & Critical & 0.40 & Open \\
User 6 & 3.64 & Critical & 0.35 & Open \\
\hline
\end{tabular}}
\caption{Health Literacy and Belief Commitment Categories. HL categories are mapped from the Total FCCHL mean: 1.00-1.99 = Functional, 2.00- 2.99 = Communicative, and 3.00-4.00 = Critical. BC categories are mapped from \(BC_{\text{total}}\): \(BC_{\text{total}} < 1.5\) = Open, \(1.5 \leq BC_{\text{total}} < 5\) = Hesitant, and \(BC_{\text{total}} \geq 5\) = Strong.}
\label{tab:hl_bc_categories}
\end{table}

\begin{figure}
    \centering
    \includegraphics[width=\linewidth]{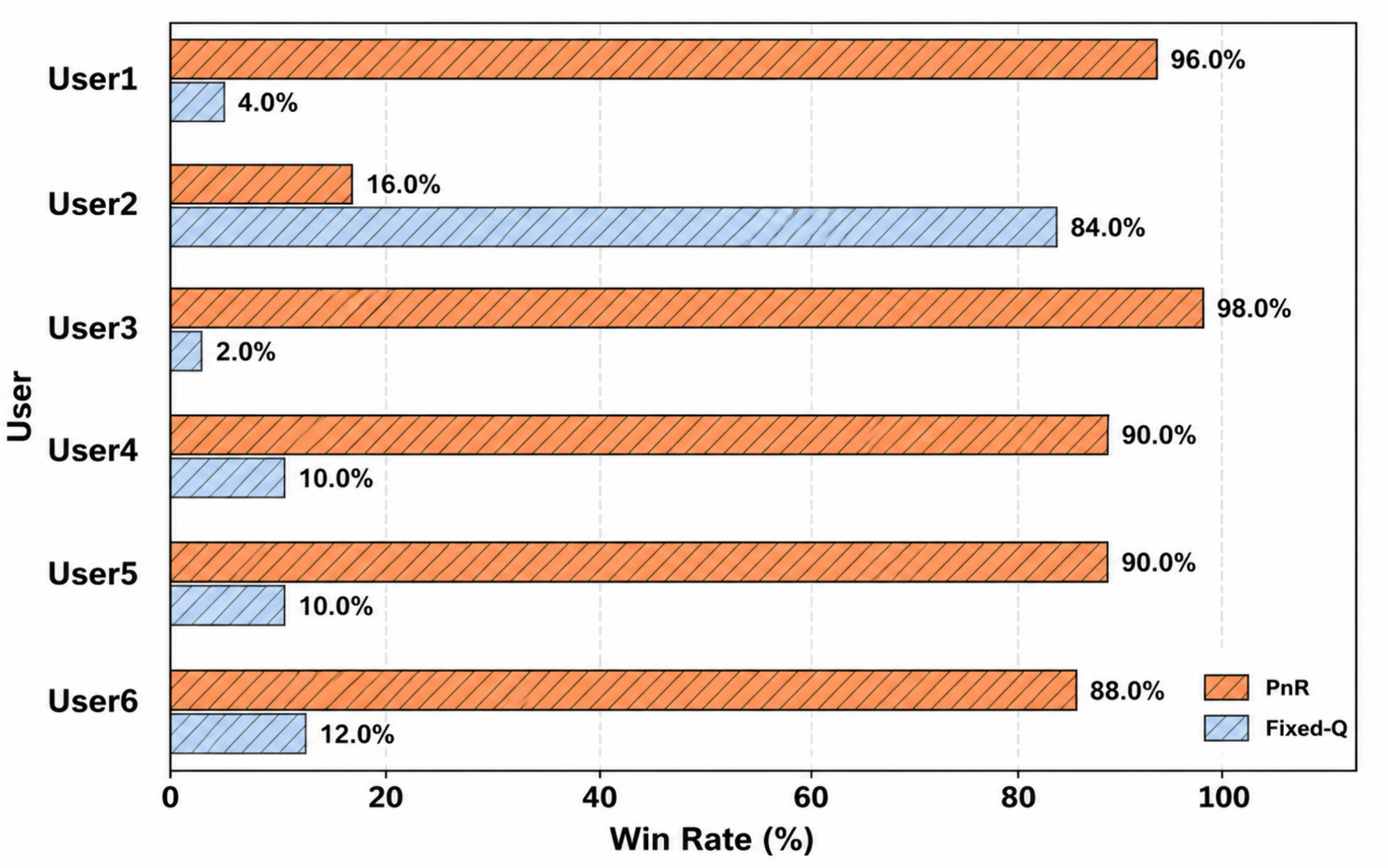}
   \caption{Human pairwise preference comparison between RO-PnR and Fixed-Q across different users. Detailed users' background are provided in Table~\ref{tab:hl_bc_categories}.}
    \label{fig:PnR_vs_FixedQ}
\end{figure}

\subsection{LLM Simulator Evaluation}
\label{Appendix: LLM Simulator Evaluation}

We recruit three PhD students in health informatics to evaluate the quality of the LLM-based user simulator along two dimensions: persona accuracy and persona consistency. Each annotator independently rates 100 sampled simulated dialogues, spanning all 9 user profiles, on a 5-point Likert scale according to the rubric in Table~\ref{tab:user-simulation-rubric}. For each annotation instance, annotators are shown the original health misinformation post, the full dialogue history, and the assigned user profile definition, including the target levels of health literacy and belief commitment. User utterances are explicitly marked in the dialogue to encourage annotators to focus on whether the simulated user's replies match the assigned profile. 
Before the formal annotation, annotators are trained with a small set of example dialogues and discuss the rating criteria to ensure a shared understanding of the two persona dimensions. After independent annotation, cases with substantial rating disagreement are reviewed in a discussion session. A health informatics expert then adjudicates the final label for unresolved cases, which is used for the final analysis.

Each annotator spent approximately three hours completing the rating task. We compensated each participant at a rate of \$20 per hour (totaling \$60 per annotator), which is above the local minimum wage and consistent with standard compensation for graduate-student annotators in NLP research. All annotators participated voluntarily and provided informed consent prior to the task.

To assess annotation reliability, we compute pairwise agreement and Scott's $\pi$ among the three annotators on both dimensions (Table~\ref{tab:persona_agreement}). Annotators show substantial agreement on persona accuracy (mean pairwise agreement $= 0.83$, multi-rater Scott's $\pi = 0.70$) and persona consistency (mean pairwise agreement $= 0.85$, multi-rater Scott's $\pi = 0.68$), indicating that the simulator can reliably instantiate the intended user profiles and maintain them across the dialogue.

\subsection{Human-judge agreement}
\label{Appendix: Human-judge agreement}
We recruit three PhD students in health informatics to assess the reliability of the LLM judge. Following \citet{thakur2025judging}, we quantify alignment between human annotations and LLM ratings using percent agreement and Scott’s $\pi$ coefficient~\cite{scott1955reliability}.
We sample  100 samples from the results, stratified across (a) the three models (Llama / Gemma / Qwen), (b) all baseline and finetuning methods, and (c) the 9 user-profile cells.
We first train the annotators on 10 practice samples to familiarize them with the task and scoring rubric. They then rate the responses independently through individual annotation links, without discussion or coordination. Afterward, we identify the cases with the largest disagreement and conduct an adjudication discussion involving the three annotators and a domain expert. The final score range for each such case is then determined based on this review.

The three human evaluators show good internal consistency (Table~\ref{tab:human_internal_agreement for AA_PG_TA}), with Scott's $\pi$ ranging from $0.63$ to $0.78$ under 0.5-point discretization, moderate to strong agreement after correcting for chance. We then use the mean of the three human ratings as a consensus and compare it against the LLM judge. The two align reasonably well (Table~\ref{tab:Human_LLM_agreement}): on the aggregated \textit{Quality} score, Scott's $\pi = 0.55$ with $73\%$ exact agreement after binning, MAE $= 0.23$, and $87\%$ of samples falling within $\pm 0.5$ points. Rank-order correlation is also substantial (Pearson $= 0.57$, Spearman $= 0.72$; Table~\ref{tab:judge_human_correlation}), indicating that the LLM judge and the human consensus rank responses similarly even when their exact scores differ slightly. Agreement is strongest on \textit{Tailored Actionability} and overall \textit{Quality}, while \textit{Personalized Grounding} is the weakest dimension across both inter-human and human-LLM comparisons. Together, these results suggest that the LLM judge is a reliable proxy for human evaluation at scale.

\subsection{Human pairwise preference}
\label{Appendix: Human pairwise preference}
Due to budget constraints, we recruited six users with varying educational backgrounds: two with middle school education, two with high school education, and two PhD students. 
This sampling strategy was informed by prior evidence showing a positive association between educational attainment and health literacy~\cite{ishikawa2008measuring}. 
Each participant spent approximately one hour completing the screening questionnaire and the pairwise preference task, and was compensated with a \$25 gift card. This rate is above the U.S. federal minimum wage and consistent with standard compensation for lay-user studies in NLP and HCI research. All participants provided informed consent prior to the study.

\begin{figure*}
    \centering
    \includegraphics[width=1\linewidth]{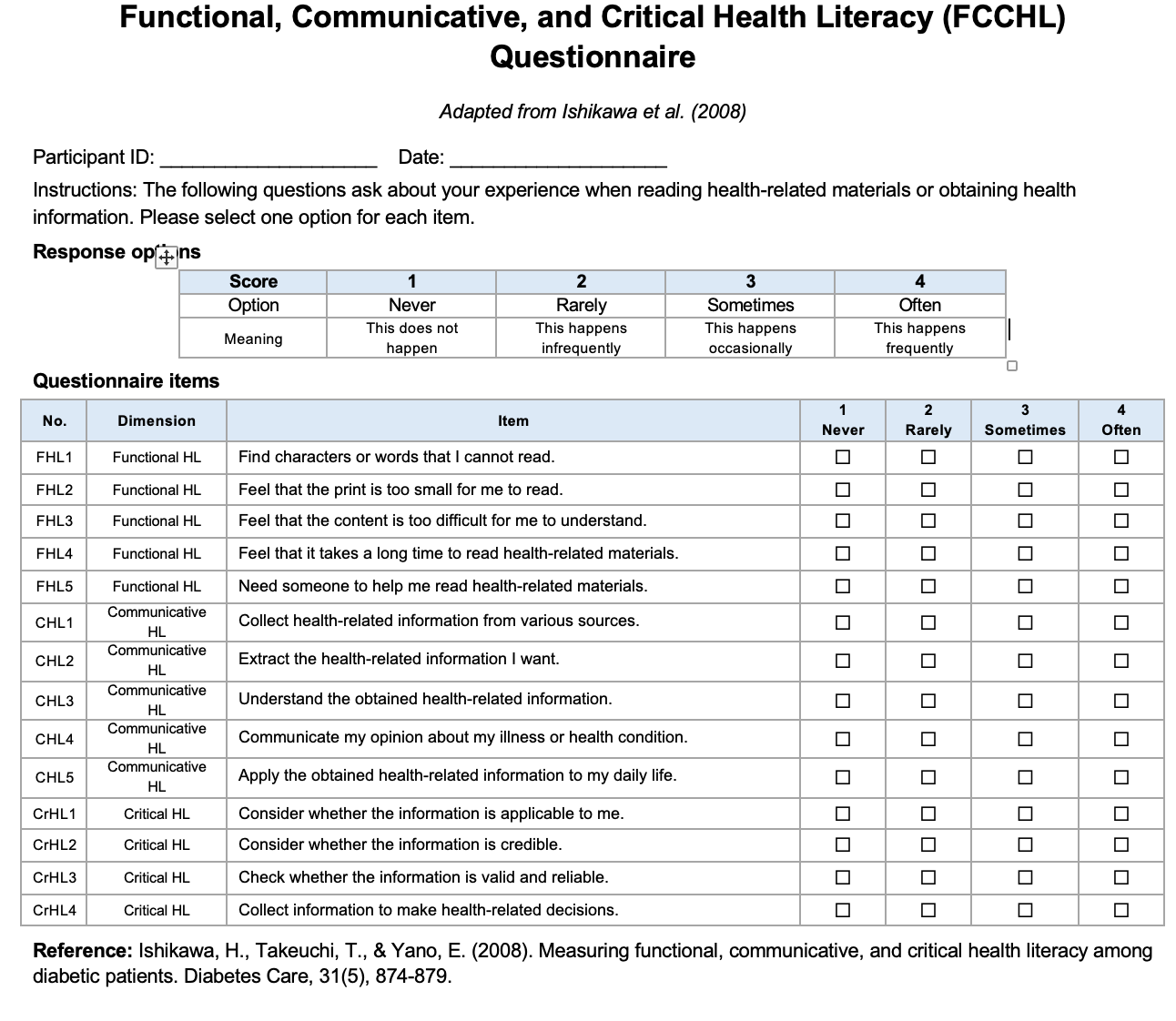}
    \caption{Questionnaire for health literacy screen}
    \label{fig:Questionnaire}
\end{figure*}

We assessed their health literacy using the Functional, Communicative, and Critical Health Literacy (FCCHL) scale~\cite{ishikawa2008measuring}.
The FCCHL scale aligns with Nutbeam’s categorization of health literacy into functional, communicative, and critical dimensions \citep{nutbeam2000health}, see questionnaire in Figure \ref{fig:Questionnaire}. Based on the participants’ FCCHL scores, we used the sample median to categorize them into lower and higher health literacy groups.

In terms of the belief commitment screening measure, given that there are no established methods available for this purpose, we define belief commitment as the joint product of how strongly a user accepts a claim and how resistant they are to revising it. Therefore, we capture both components using three 0--10 slider items for each of the three target misinformation claims, as shown in Table~\ref{tab:belief_commitment_items}.
For each claim \(i\), we compute a 0-10 commitment score that gates acceptance \((B_i)\) by a strength multiplier formed from certainty \((C_i)\) and the complement of openness \((O_i)\):

\[
\text{commitment}_i
=
B_i \times \frac{C_i + (10 - O_i)}{20}.
\]

If the user does not accept the claim, meaning \(B_i\) is near 0, then the commitment score is near 0 regardless of certainty. If the user accepts the claim, is highly certain, and is unwilling to revise their view, the commitment score approaches 10. The per-user belief commitment score is the mean across the three claims:

\[
BC_{\text{total}} = \frac{1}{3}\sum_{i=1}^{3}\text{commitment}_i
\]
where \(BC_{\text{total}}\) ranges from 0 to 10.

The final screening results are shown in Table~\ref{tab:hl_bc_categories}. Among the six users, three were classified as having a critical HL and open BC background, one as having a communicative HL and open BC background, one as having a communicative HL and hesitant BC background, and one as having a functional HL and strong-believer BC background. 

We sample 50 pairs from one of the best baselines (Fixed-Q) and RO-PnR (50*2). Each pair uses the same misinformation post and the same user profile, with the RO-PnR response and the baseline response shown side-by-side, anonymized, in randomized order. 
We conduct a blind pairwise preference evaluation on 50 health misinformation posts using six users with diverse health literacy levels and belief commitment profiles. For each post, users see only the health misinformation post and two anonymous counterspeech responses, without knowing whether the responses are generated by \textit{RO-PnR} or \textit{Fixed-Q}. 

\section{Use of AI Assistants}
We used AI-assisted tools for support with code development and language refinement. 
All core ideas, methodological decisions, analyses, interpretations, and conclusions were developed by the authors. 


\begin{figure*}[ht]
    \centering
    \begin{outerbox}
        
        \begin{casebox}{darkblue}{lightblue}
        \raggedright
            \textbf{System Prompt (\texttt{Roleplay / Health Misinformation Persona}):}

            You are simulating a realistic Reddit user who previously posted a health misinformation claim and is now chatting with an assistant. Stay strictly in the \textbf{user} role. Do not act as an expert, educator, or assistant. Respond naturally as a Reddit user with personal opinions and emotions. Keep replies short (1--2 sentences).
            \vspace{6pt} \hrule \vspace{6pt}
            
            \textbf{Current User State:}

            \texttt{\{State.to\_dict()\}}
            \vspace{6pt} \hrule \vspace{6pt}
            
            \textbf{User State Dimension Guide:}

            \texttt{\{guide\_text\}}

            Interpret \texttt{CURRENT\_USER\_STATE} using the guide above and stay aligned with it.
            \vspace{6pt} \hrule \vspace{6pt}
            
            \textbf{Conversation So Far:}

            \texttt{\{history\_text if history\_text else '(no prior conversation)'\}}
            \vspace{6pt} \hrule \vspace{6pt}
            
            \textbf{Assistant Message:}

            \texttt{\{question\}}
            \vspace{6pt} \hrule \vspace{6pt}
            
            \textbf{Generation Instruction:}

            Generate exactly one realistic Reddit-style user reply.
        \end{casebox}

        \vspace{-2pt}

        \begin{casebox}{darkblue}{lightblue}
        \raggedright
            \textbf{User Prompt / Output Constraints (\texttt{JSON Reply Formatting}):}

            Guidelines:
            \begin{itemize}[leftmargin=14pt, topsep=2pt, itemsep=2pt]
                \item Speak casually like a real Reddit user.
                \item Do not summarize the belief state.
                \item Do not explain medical facts unless the user would naturally know them.
                \item Do not ask multiple questions.
                \item Keep the reply brief.
                \item Remain in character as a user throughout the response and follow the behavior rules carefully.
            \end{itemize}
            \vspace{4pt} \hrule \vspace{6pt}
            
            \textbf{Behavior Rules:}

            \texttt{\{behavior\_rules\}}
            \vspace{6pt} \hrule \vspace{6pt}
            

        \end{casebox}

    \end{outerbox}
    \caption{System prompt and user prompt for realistic health misinformation persona simulation.}
    \label{fig:user-prompt}
\end{figure*}

\begin{figure*}[ht]
    \centering
    \begin{outerbox}
        \begin{casebox}{darkblue}{lightblue}
        \raggedright
            \textbf{System Prompt:}

            You are a professional fact-checking judge. Your task is to determine whether the given content is factually accurate.

            Use reliable external sources, together with your own knowledge, to verify the response.

            \medskip
            \textbf{Response:}

            ``\{final\_response\}''

            \medskip
            \textbf{Evaluation Instructions:}
            \begin{itemize}
                \item Output \texttt{Label: 1} if the response is factually correct and does not contain false or misleading claims.
                \item Output \texttt{Label: 0} if the response contains any false, misleading, or unsupported health-related claim.
                \item Output \texttt{Label: UNCERTAIN} if the factual accuracy cannot be determined after verification.
                \item Provide a brief reason for your judgment in 1--3 sentences.
            \end{itemize}

            \medskip
            \textbf{Output Format:}

            \texttt{Label:} \texttt{0}, \texttt{1}, or \texttt{UNCERTAIN}

            \texttt{Reason:} brief explanation of the judgment
        \end{casebox}
    \end{outerbox}
    \caption{Prompt for evaluating the Factual Error.}
    \label{fig:fact-checking-judge-prompt}
\end{figure*}


\begin{figure*}[ht]
    \centering
    \begin{outerbox}
        
        \begin{casebox}{darkyellow}{lightyellow}
        \raggedright
            \textbf{Audience Alignment}

            Whether the agent frames the correction in a way that is appropriate for the user’s inferred health literacy and belief commitment.
            \vspace{6pt} \hrule \vspace{6pt}
            
            \textbf{Score Range 1.0--2.0:}

            unclear, misleading, badly framed, or likely to provoke rejection.
            \vspace{4pt}
            
            \textbf{Score Range 2.1--3.2:}

            directionally appropriate, but weak, generic, indirect, or poorly matched to the user’s literacy or belief commitment.

            \vspace{4pt}
            
            \textbf{Score Range 3.3--3.8:}

            reasonably clear and credible, with some fit to the user’s stance, literacy, and comprehension.

            \vspace{4pt}
            
            \textbf{Score Range 3.9--4.4:}

            clearly frames the correction in a way that is well adapted to the user’s stance, literacy, and belief commitment.

            \vspace{4pt}
            
            \textbf{Score Range 4.5--5.0:}

           exceptionally clear, credible, well-framed, and highly responsive to the user’s literacy, comprehension, and belief commitment.
        \end{casebox}

        \vspace{-2pt}

        \begin{casebox}{darkyellow}{lightyellow}
        \raggedright
            \textbf{Personalized Grounding}

            Whether the agent gives useful, user-comprehensible evidence, mechanisms, or reasoning behind the correction.
            \vspace{6pt} \hrule \vspace{6pt}
            
            \textbf{Score Range 1.0--2.0:}

            missing, inaccurate, vague, or mismatched to the user’s comprehension.
            \vspace{4pt}
            
            \textbf{Score Range 2.1--3.2:}

            includes some reason or evidence, but it is shallow, generic, weakly connected, or not well matched to the user’s understanding.
            \vspace{4pt}
            
            \textbf{Score Range 3.3--3.8:}

            provides useful reason, evidence, or mechanism that is moderately tailored to the user’s comprehension.
            \vspace{4pt}
            
            \textbf{Score Range 3.9--4.4:}

            provides clear, relevant, and understandable evidence or reasoning that is well matched to the user’s level and stance.
            \vspace{4pt}
            
            \textbf{Score Range 4.5--5.0:}

            provides especially sharp, credible, and well-translated evidence, mechanisms, or reasoning that strongly improves the user’s understanding.
        \end{casebox}

        \vspace{-2pt}

        \begin{casebox}{darkyellow}{lightyellow}
        \raggedright
            \textbf{Tailored Actionability}

            Whether the agent provides safe, concrete, and user-appropriate actionable guidance when needed.
            \vspace{6pt} \hrule \vspace{6pt}
            
            \textbf{Score Range 1.0--2.0:}

            no meaningful next step, unsafe, unrealistic, not actionable, or mismatched to the user’s situation.
            \vspace{4pt}
            
            \textbf{Score Range 2.1--3.2:}

            offers some direction, but it is vague, generic, awkward, or only weakly matched to the user’s needs.
            \vspace{4pt}
            
            \textbf{Score Range 3.3--3.8:}

            provides a reasonable, safe next step that the user could follow, with some attention to the user’s situation.
            \vspace{4pt}
            
            \textbf{Score Range 3.9--4.4:}

            provides clear, practical, safe, and user-appropriate guidance that supports informed decision-making.
            \vspace{4pt}
            
            \textbf{Score Range 4.5--5.0:}

            provides unusually strong, natural, highly actionable, and well-tailored guidance, such as appropriate next steps or trusted information sources that fit the user’s needs.
        \end{casebox}

    \end{outerbox}
    \caption{Evaluation rubric for Audience Alignment, Personalized Grounding, and Tailored Actionability.}
    \label{fig:twn-rubric}
\end{figure*}



\begin{figure*}[ht]
    \centering
    \begin{outerbox}
        \begin{casebox}{darkblue}{lightblue}
        \raggedright
            \textbf{System Prompt:}

            You are a health information assistant engaged in a multi-turn clarification interaction about possible health misinformation. At the current turn, decide whether to ask one additional clarification question or to stop asking questions and produce the final response. Do not assume any hidden user profile. Base your decision only on the health misinformation post, the dialogue history, and the latest user reply.
        \end{casebox}
    \end{outerbox}
    \caption{System prompt for the Reactive baseline.}
    \label{fig:rpr-system-prompt}
\end{figure*}

\begin{figure*}[ht]
    \centering
    \begin{outerbox}
        \begin{casebox}{darkblue}{lightblue}
        \raggedright
            \textbf{Health Misinformation Post:}

            \texttt{\{hm\_post\}}
            \vspace{4pt} \hrule \vspace{4pt}
            
            \textbf{Dialogue History:}

            \texttt{\{history\}}
            \vspace{4pt} \hrule \vspace{4pt}
            
            \textbf{Latest User Reply:}

            \texttt{\{latest\_user\_reply\}}
            \vspace{4pt} \hrule \vspace{4pt}
            
            \textbf{Task:}

            Decide between the following two actions:
            \begin{itemize}[leftmargin=14pt, topsep=2pt, itemsep=2pt]
                \item \textbf{probe}: ask one additional clarification question if the user’s concern, evidence basis, or interpretation remains unclear and one more question would likely help improve the final response.
                \item \textbf{respond}: stop asking questions if the current dialogue already provides enough information to produce an effective final response.
            \end{itemize}
            \vspace{4pt} \hrule \vspace{4pt}
            
            \textbf{Decision Guidance:}

            \begin{itemize}[leftmargin=14pt, topsep=2pt, itemsep=2pt]
                \item Choose \textbf{probe} when an important uncertainty remains unresolved.
                \item Choose \textbf{respond} when the user’s concern is already sufficiently clear and another question would likely be redundant.
                \item Ask at most one question.
                \item Do not repeat or paraphrase a previous question.
                \item If you choose \textbf{probe}, the question must be brief, natural, and directly useful for improving the final counterspeech.
                \item If you choose \textbf{respond}, do not ask any additional question.
                \item If you choose \textbf{respond}, the counterspeech should follow the \textbf{AA/PG/TA} principle:
                  \begin{itemize}
        \item \textbf{AA (Audience Alignment):} Frame the correction appropriately for the user's inferred health literacy and belief commitment.
        \item \textbf{PG (Personalized Grounding):} Provide clear, user-comprehensible evidence or reasoning to support the correction.
        \item \textbf{TA (Tailored Actionability):} Offer safe, concrete next steps suited to the user when needed.
    \end{itemize}
                \item Use only the visible conversation context. Do not mention or assume any hidden user state.
            \end{itemize}
            \vspace{4pt} \hrule \vspace{4pt}
            
            \textbf{Output Constraints:}

            \begin{itemize}[leftmargin=14pt, topsep=2pt, itemsep=2pt]
                \item Return only the required tagged fields.
                \item Do not include explanations, reasoning, or extra text outside the tags.
                \item If the decision is \textbf{probe}, fill the \texttt{<question>} field and leave \texttt{<response>} empty.
                \item If the decision is \textbf{respond}, fill the \texttt{<response>} field and leave \texttt{<question>} empty.
            \end{itemize}




        \end{casebox}
    \end{outerbox}
    \caption{Prompt for the Reactive baseline.}
    \label{fig:rpr-decision-instructions}
\end{figure*}


\begin{figure*}[ht]
    \centering
    \begin{outerbox}
        \begin{casebox}{darkblue}{lightblue}
        \raggedright
            \textbf{System Prompt:}

            You are a health information assistant engaged in a multi-turn clarification interaction about possible health misinformation. At the current turn, infer the user’s likely state from the observed dialogue and decide whether your confidence is high enough to stop asking questions and produce the final response. Do not assume any hidden user profile explicitly. Base your decision only on the health misinformation post, the dialogue history, and the latest user reply.
        \end{casebox}
    \end{outerbox}
    \caption{System prompt for the Confidence baseline.}
    \label{fig:cgp-system-prompt}
\end{figure*}

\begin{figure*}[ht]
    \centering
    \begin{outerbox}
        \begin{casebox}{darkblue}{lightblue}
        \raggedright
            \textbf{Health Misinformation Post:}

            \texttt{\{hm\_post\}}
            \vspace{6pt} \hrule \vspace{6pt}
            
            \textbf{Dialogue History:}

            \texttt{\{history\}}
            \vspace{6pt} \hrule \vspace{6pt}
            
            \textbf{Latest User Reply:}

            \texttt{\{latest\_user\_reply\}}
            \vspace{6pt} \hrule \vspace{6pt}
            
            \textbf{Task:}

            First infer the user’s likely state from the observed dialogue. Then decide between the following two actions:
            \begin{itemize}[leftmargin=14pt, topsep=2pt, itemsep=2pt]
                \item \textbf{probe}: ask one additional clarification question if your confidence in the inferred user state is still too low and one more question would help reduce uncertainty.
                \item \textbf{respond}: stop asking questions if your confidence in the inferred user state is high enough to produce an effective final counterspeech response.
            \end{itemize}
            \vspace{4pt} \hrule \vspace{6pt}
            

            
            \textbf{Decision Guidance:}

            \begin{itemize}[leftmargin=14pt, topsep=2pt, itemsep=2pt]
                \item Choose \textbf{probe} when your estimate of the user’s state is still uncertain.
                \item Choose \textbf{respond} when the user’s likely state is sufficiently clear and another question is unlikely to substantially improve the final counterspeech.
                \item Ask at most one question.
                \item Do not repeat or paraphrase a previous question.
                \item If you choose \textbf{probe}, the question must be brief, natural, and directly useful for reducing uncertainty about the user’s likely state or concern.
                \item If you choose \textbf{respond}, do not ask any additional question.
                \item If you choose \textbf{respond}, the counterspeech should follow the \textbf{AA/PG/TA} principle:
                  \begin{itemize}
        \item \textbf{AA (Audience Alignment):} Frame the correction appropriately for the user's inferred health literacy and belief commitment.
        \item \textbf{PG (Personalized Grounding):} Provide clear, user-comprehensible evidence or reasoning to support the correction.
        \item \textbf{TA (Tailored Actionability):} Offer safe, concrete next steps suited to the user when needed.
    \end{itemize}
                \item Use only the visible conversation context. Do not mention or assume any hidden user state.
            \end{itemize}
            \vspace{4pt} \hrule \vspace{6pt}
            
            \textbf{Confidence Threshold Rule:}

            Use an internal confidence threshold for deciding whether to stop probing:
            \begin{itemize}[leftmargin=14pt, topsep=2pt, itemsep=2pt]
                \item If confidence in the inferred user state is high, choose \textbf{respond}.
                \item If confidence in the inferred user state is not yet high, choose \textbf{probe}.
            \end{itemize}
            Do not output the threshold value or your reasoning.
            \vspace{4pt} \hrule \vspace{6pt}
            
            \textbf{Output Constraints:}

            \begin{itemize}[leftmargin=14pt, topsep=2pt, itemsep=2pt]
                \item Return only the required tagged fields.
                \item Do not include explanations, reasoning, confidence scores, or extra text outside the tags.
                \item If the decision is \textbf{probe}, fill the \texttt{<question>} field and leave \texttt{<response>} empty.
                \item If the decision is \textbf{respond}, fill the \texttt{<response>} field and leave \texttt{<question>} empty.
            \end{itemize}
            

        \end{casebox}
    \end{outerbox}
    \caption{Prompt for the Confidence baseline.}
    \label{fig:cgp-decision-instructions}
\end{figure*}

\end{document}